\documentclass[a4paper,fleqn]{cas-sc}

\usepackage{graphicx}
\usepackage{dcolumn}
\usepackage{bm}
\usepackage{booktabs}
\usepackage{hyperref}
\usepackage{algorithm}
\usepackage{algpseudocode}
\usepackage{float} 
\usepackage{color} 
\usepackage{tabularx}
\usepackage{subcaption}
\usepackage{tikz}
\usepackage{placeins}

\usepackage[numbers]{natbib}

\makeatletter
\ifx\those@algorithms\undefined
  \newcount\those@algorithms
\fi
\makeatother

\makeatletter
\newcommand{\removelatexerror}{\let\@latex@error\@gobble}
\makeatother

\extrafloats{100}

\begin{document}
\let\WriteBookmarks\relax
\def\floatpagepagefraction{0.8} 
\def\textpagefraction{0.01}

\shorttitle{PoreDiT: A Scalable Generative Model for Digital Rock Reconstruction}
\shortauthors{Yizhuo Huang et al.}

\title [mode = title]{PoreDiT: A Scalable Generative Model for Large-Scale Digital Rock Reconstruction}

\author[1]{Yizhuo Huang}

\author[2]{Baoquan Sun}

\author[3]{Haibo Huang}
\cormark[1]
\ead{huanghb@ustc.edu.cn} 

\cortext[cor1]{Corresponding author} 

\affiliation[1]{organization={College of Intelligent Robotics and Advanced Manufacturing, FuDan University},
            city={Shanghai},
            country={China}}

\affiliation[2]{organization={State Key Laboratory of Shale Oil and Gas Enrichment Mechanisms and Efficient Development},
            city={Dongying},
            state={Shandong},
            country={China}}

\affiliation[3]{organization={Department of Modern Mechanics, University of Science and Technology of China},
            addressline={Hefei 230026},
            city={Hefei},
            state={Anhui},
            country={China}}

\begin{abstract}
This manuscript presents PoreDiT, a novel generative model designed for high-efficiency digital rock reconstruction at gigavoxel scales. Addressing the significant challenges in digital rock physics (DRP), particularly the trade-off between resolution and field-of-view (FOV), and the computational bottlenecks associated with traditional deep learning architectures, PoreDiT leverages a three-dimensional (3D) Swin Transformer to break through these limitations. By directly predicting the binary probability field of pore spaces instead of grayscale intensities, the model preserves key topological features critical for pore-scale fluid flow and transport simulations. This approach enhances computational efficiency, enabling the generation of ultra-large-scale ($1024^3$ voxels) digital rock samples on consumer-grade hardware. Furthermore, PoreDiT achieves physical fidelity comparable to previous state-of-the-art methods, including accurate porosity, pore-scale permeability, and Euler characteristics. The model's ability to scale efficiently opens new avenues for large-domain hydrodynamic simulations and provides practical solutions for researchers in pore-scale fluid mechanics, reservoir characterization, and carbon sequestration.
\end{abstract}

\begin{highlights}
\item PoreDiT, a 3D Swin Transformer model, enables gigavoxel-scale digital rock reconstruction on consumer-grade GPUs.
\item Directly predicts binary probability fields to preserve micro-scale topological features and avoid grayscale artifacts.
\item Achieves true 3D generation of $1024^3$ voxels, overcoming the field-of-view and resolution trade-off in DRP.
\item Validated against state-of-the-art methods, demonstrating high physical fidelity in porosity, permeability, and connectivity.
\end{highlights}


\maketitle

\section{Introduction}

Understanding the physics of fluid transport in porous media is a 
cornerstone of diverse scientific and engineering disciplines. 
It governs critical processes such as multiphase flow in hydrocarbon 
reservoirs, CO$_2$ plume migration in carbon sequestration, 
and groundwater contaminant dispersion \citep{blunt2013pore, bear2013dynamics}. 
To elucidate the complex displacement mechanisms—such as spontaneous 
imbibition and capillary hysteresis—within these opaque systems, 
Digital Rock Physics (DRP) has emerged as an indispensable paradigm. 
By digitizing the pore structure, researchers can perform direct 
pore-scale simulations (e.g., Lattice Boltzmann Method, 
Pore Network Modeling) to investigate how micro-topology 
influences macroscopic transport properties. 
Recent studies have demonstrated 
the efficacy of DRP in revealing mechanisms of counter-current 
spontaneous imbibition in fractured media 
\citep{hu2025pore} and modeling 
nonlinear gas flow in organic matter \citep{liu2025pore}. 

However, the fidelity of any pore-scale hydrodynamic simulation is strictly bounded by the geometric accuracy and the scale of the digital rock model.
A fundamental computational challenge lies in the trade-off between spatial resolution and Field-of-View (FOV). 
Resolving the fine-scale pore throats requires micro-computed tomography (micro-CT) imaging at high magnification, 
which inevitably restricts the sample volume. 
As CT data is intrinsically three-dimensional, the computational grid grows cubically with image dimensions, 
causing numerical reconstructions and subsequent flow simulations to rapidly become compute- and memory-bound \cite{lopes2025enabling}. While recent advancements have sought to alleviate these bottlenecks through efficient memory management and architecture optimization in multiscale modeling \cite{zhu2025data}, 
memory usage continues to be a severe limitation for analyzing giga-voxel volumes on accessible, consumer-grade hardware. Consequently, digital domains often fail to meet the Representative Elementary Volume (REV) requirement, introducing significant uncertainty into hydrodynamic simulations \citep{bultreys2016imaging}.

To bridge this gap, efficient data-driven digital reconstruction methods have been rapidly developed to synthesize large-scale, 
high-fidelity microstructures that are statistically equivalent to real materials \cite{ge20263d, chen2025novel}. Historically, stochastic methods such as Gaussian Random Fields (GRF) and Simulated Annealing (SA) were widely used \citep{torquato1993chord, hazlett1997statistical}. While computationally efficient, these methods often struggle to reproduce the complex, non-local connectivity patterns characteristic of natural porous systems \citep{zhu2019challenges}. The advent of Deep Learning shifted the paradigm towards Generative Adversarial Networks (GANs). \cite{mosser2017reconstruction} pioneered the use of DCGANs for pore structure synthesis, and subsequent variations like WGAN-GP \citep{zha2020reconstruction} and Progressive Growing GANs \citep{you20213d} improved training stability. Despite these advances, GAN-based approaches face inherent challenges in ``mode collapse'' and often rely on two-dimensional(2D) or pseudo-three-dimensional(3D) convolutions that limit their ability to capture long-range 3D topological dependencies essential for fluid percolation. More recently, Diffusion Models have shown superior distribution matching capabilities, but current implementations often rely on memory-intensive 3D U-Net architectures \citep{zheng2022rockgpt}, which restrict the generation size ($256^3$ voxels) and hinder the simulation of large-scale flow heterogeneity.

To address these limitations and provide massive, high-fidelity geometric boundary conditions for Computational Fluid Dynamics (CFD) simulations, we propose \textbf{PoreDiT} (Pore-scale Diffusion Transformer). Our approach introduces three key methodological and innovative achievements to the field of microstructure reconstruction:
\begin{enumerate}
    \item \textbf{Architectural Innovation:} We introduce the first Voxel-based Transformer architecture for porous media diffusion, replacing the memory-intensive 3D U-Net and significantly reducing computational complexity.
    \item \textbf{Gigavoxel Scalability:} We break the hardware barrier to achieve ultra-large-scale ($1024^3$ voxels) reconstruction on consumer-grade Graphics Processing Units(GPUs)(e.g., RTX 4090), making high-performance computing accessible for local workstations
    \item \textbf{Physical Fidelity:} Extensive validation demonstrates that PoreDiT achieves physical fidelity comparable to existing state-of-the-art methods in morphological and topological fidelity, including porosity, permeability, and Euler characteristics.
\end{enumerate}

By bridging the gap between deep generative modeling and fluid 
dynamics, PoreDiT fundamentally advances the capabilities of 
computational pore-scale flow studies. Specifically, 
it provides the massive, REV-compliant geometric boundary 
conditions necessary for investigating scale-dependent hydrodynamic 
phenomena, ensuring that the synthesized domains are not merely 
visually realistic but \textbf{hydraulically equivalent}. 
This practical utility is rigorously validated through Lattice 
Boltzmann Method (LBM) simulations, where PoreDiT-generated 
samples yield permeability distributions and flow correlations 
that tightly match laboratory measurements, thereby validating 
the model as a reliable tool for predictive hydrodynamics.

\label{intro}

\section{Related Works}

\subsection{Stochastic Reconstruction and GAN-based Approaches}

The reconstruction of porous media has historically relied on \textit{stochastic methods}. Early seminal works utilizing Gaussian Random Fields (GRF) \citep{torquato1993chord} and Simulated Annealing (SA) \citep{hazlett1997statistical} established the foundational framework for statistical morphology modeling. However, as comprehensively reviewed by Zhu et al.\cite{zhu2019challenges}, these conventional approaches face significant limitations: GRF methods often struggle to reproduce non-Gaussian long-range connectivity, while optimization-based techniques like SA are computationally prohibitive and prone to local optima, hindering their scalability to large-scale volumes.

The advent of deep learning shifted the paradigm towards generative modeling, with \textit{Generative Adversarial Networks (GANs)} leading the initial wave. Mosser et al. \cite{mosser2017reconstruction} pioneered the application of DCGANs to implicitly learn the probability distribution of pore structures. Despite this breakthrough, standard GANs are notoriously unstable and prone to \textit{mode collapse} \citep{zha2020reconstruction}. Although subsequent variations such as WGAN-GP \citep{zha2020reconstruction} and Progressive Growing GANs \citep{you20213d} were introduced to improve training stability and image resolution, and Pyramid WGANs \citep{zhu2024generation} attempted to capture multi-scale features, these methods fundamentally rely on Convolutional Neural Networks (CNNs). The inherent local receptive field of convolution operations limits their ability to capture global topological dependencies, which are critical for the physical fidelity of complex porous media.

\subsection{Transformers in Porous Media}

Following the paradigm shift introduced by \textit{Vision Transformers (ViT)} \citep{dosovitskiy2020image}, which demonstrated the superiority of self-attention in capturing global dependencies, researchers have begun exploring their potential in digital rock physics. RockGPT \citep{zheng2022rockgpt} represents a pioneering effort, utilizing a Generative Pre-trained Transformer (GPT)-based architecture to autoregressively predict 3D structures. However, this approach relies on Vector Quantized-Variational Autoencoder(VQ-VAE) to compress pore structures into discrete codebooks. This ``two-stage'' strategy inevitably incurs a loss of high-frequency details, while the sequential nature of autoregressive generation limits inference speed. Currently, high-fidelity reconstruction using Transformers directly on voxel representations—without pre-trained latent compression—remains largely unexplored.

\subsection{Diffusion Models and the Proposed PoreDiT}

\textit{Diffusion Probabilistic Models (DPMs)} have recently emerged as a powerful tool in scientific computing. Initial applications focused on super-resolution enhancement \citep{ma2023enhancing} or inverse material design \citep{vlassis2023denoising}. Notably, Park et al.\cite{park2024inverse} demonstrated that diffusion models could outperform GANs by over three orders of magnitude in structural validity for zeolite generation.

Most recently, Naiff et al.\cite{naiff2025controlled} advanced the state-of-the-art (SOTA) by employing a Latent Diffusion Model with a 3D U-Net backbone. While achieving high fidelity, their approach faces significant \textit{computational bottlenecks} due to the cubic complexity of 3D convolutions and the heavy reliance on high-end hardware (e.g., NVIDIA A100s), typically constraining generation to volumes of $256^3$.

To address these limitations, we introduce \textbf{PoreDiT} (Pore-scale Diffusion Transformer). Distinct from the latent space approach of Naiff et al.\cite{naiff2025controlled}, PoreDiT circumvents the need for a separate VAE compression stage. Instead, it operates on \textit{patchified voxel embeddings}, learning the binary probability field of the pore space directly. Our design philosophy aligns with the manifold hypothesis presented in the concurrent work \textit{JiT} by Li et al.\cite{li2025back}. Empirical results suggest that this strategy of directly predicting clean data ($x_0$) rather than noise enables efficient projection onto the physical manifold, allowing PoreDiT to scale to gigavoxel ($1024^3$) resolutions on consumer-grade hardware (e.g., RTX 4090).

\section{Methodology}

\begin{figure}[pos=htbp]
\centering
\includegraphics[width=0.99\textwidth]{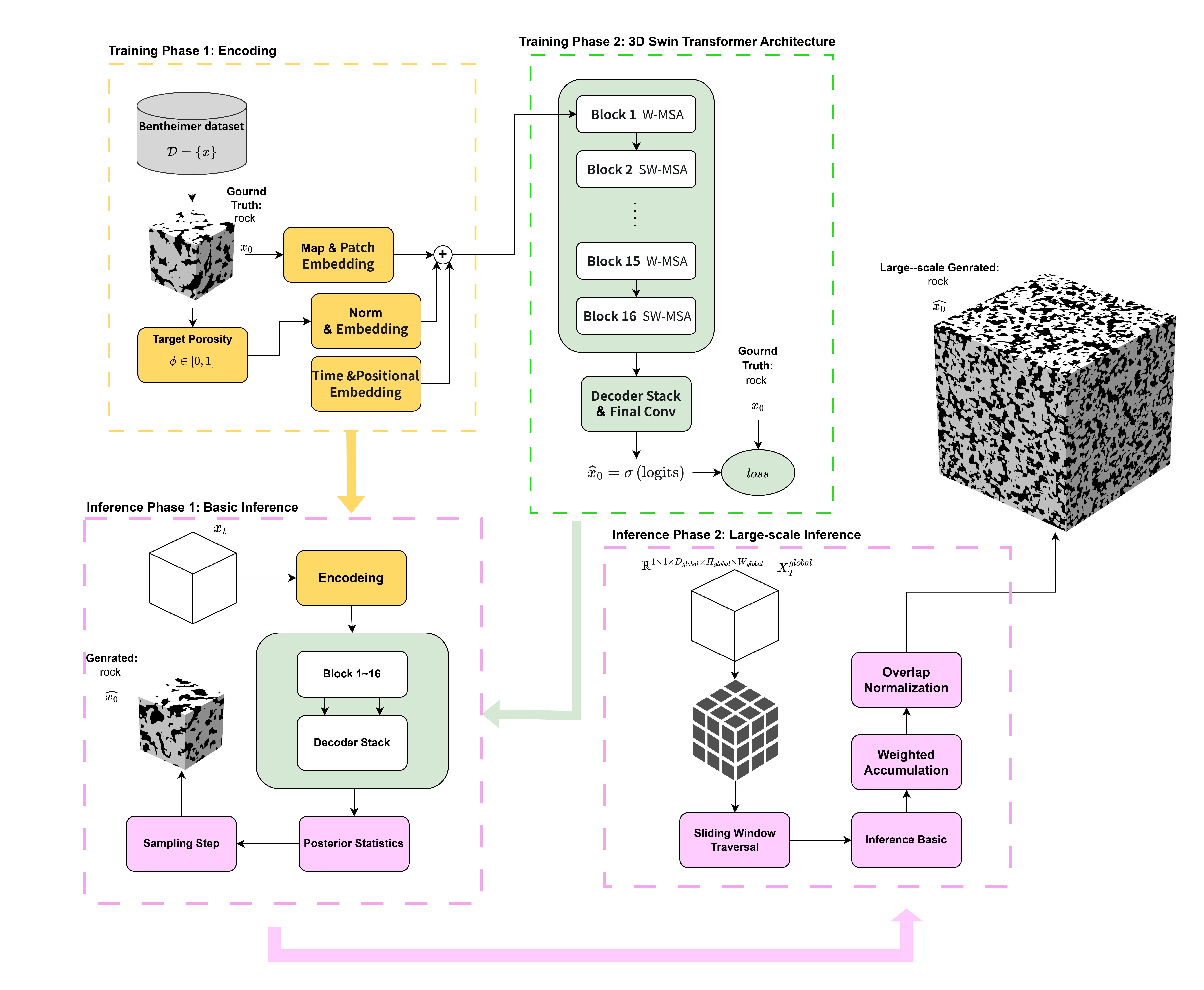}
\caption{\textbf{Overview of the PoreDiT framework and inference mechanism.} The diagram illustrates the complete workflow from microscopic training to large-scale generation. (A) \textbf{Probability Field Learning:} The model predicts the conditional probability of pore phase existence, quantifying aleatoric uncertainty. (B) \textbf{3D Patch Embedding:} Input volumes are partitioned into patch sequences to optimize memory efficiency while capturing global topology via Swin Transformers. (C) \textbf{Large-scale Inference:} A Gaussian-weighted overlap-tile strategy enables seamless gigavoxel reconstruction.}
\label{fig:framework} 
\end{figure}

As illustrated in Figure \ref{fig:framework}, we propose \textbf{PoreDiT} (Pore-scale Diffusion Transformer), a unified generative framework designed to achieve physically rigorous reconstruction at the gigavoxel scale. Rather than treating porous media generation as a standard image synthesis task, our methodology explicitly addresses the discrete nature of rock physics through a probabilistic lens. 

To simultaneously secure \textit{structural fidelity} and \textit{computational scalability}, PoreDiT integrates three strategic mechanisms: 
(1) \textbf{Reformulating generation as probability field evolution.} Unlike traditional methods that regress pseudo-grayscale density, our model learns the conditional likelihood of the pore phase. This probabilistic formulation preserves sub-voxel connectivity information and mitigates the topological distortions caused by the "blur-then-truncate" artifacts common in direct regression; 
(2) \textbf{Leveraging isotropic 3D Swin Transformers.} By replacing local convolutions with self-attention mechanisms, we capture long-range topological dependencies without the aggressive downsampling inherent in U-Net architectures, thereby ensuring the integrity of narrow pore throats; 
and (3) \textbf{Implementing a patch-based latent strategy.} This approach decouples memory consumption from generation volume, allowing for the seamless extrapolation of massive, physically homogeneous digital rock cores on consumer-grade hardware. 

\subsection{Physical Representation and Tokenization }

\begin{figure}[pos=htbp]
\centering
\includegraphics[width=0.8\textwidth]{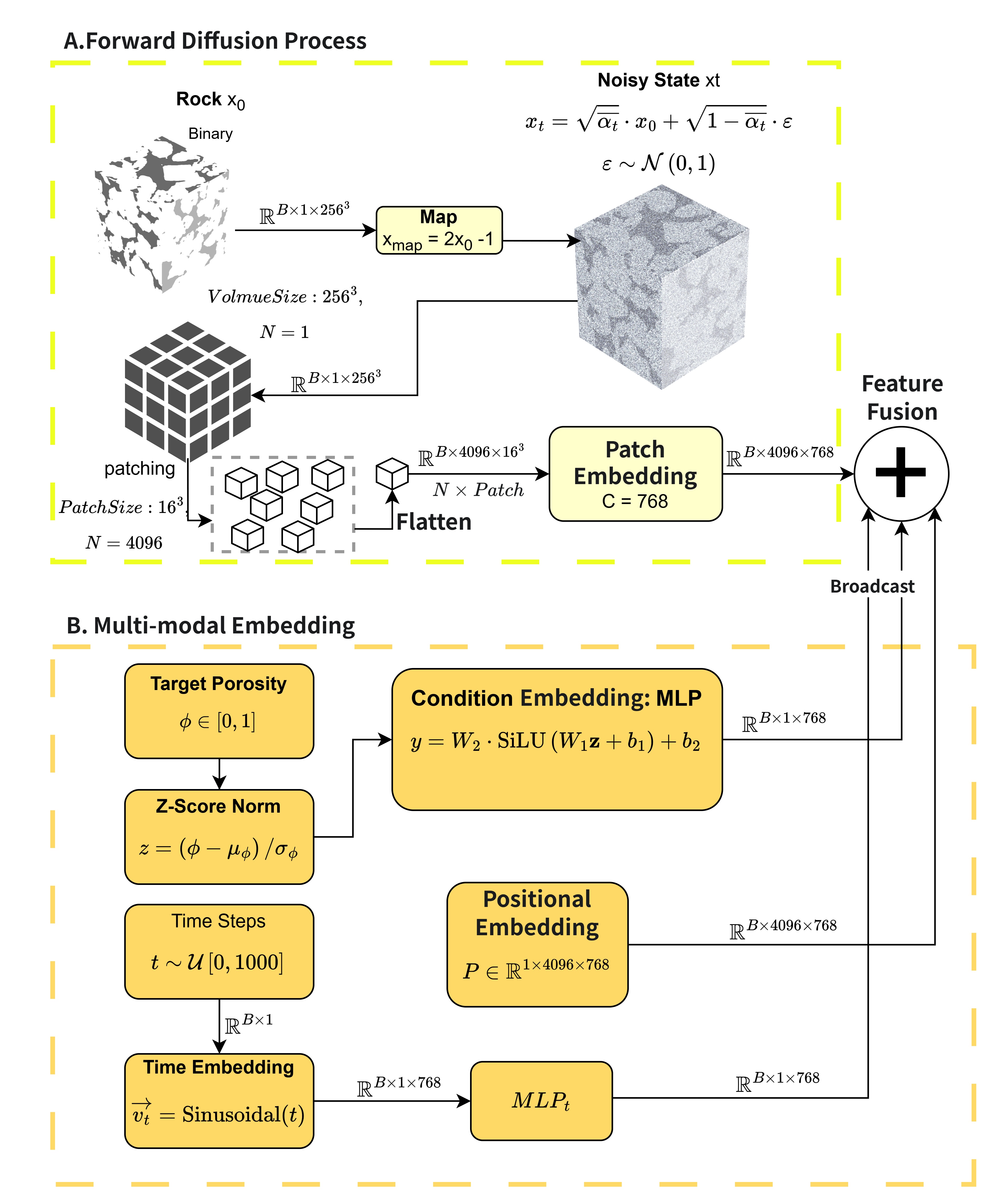} 
\caption{Schematic of the Forward Diffusion Process and Multi-modal Embedding Framework.}
\label{fig:my_structur} 

\begin{flushleft}
{\footnotesize
(A) Forward Diffusion Process: The binary rock model $x_0$ is mapped to
a continuous space and perturbed with Gaussian noise to 
generate the noisy state $x_t$. 
To adapt to the Transformer architecture, the volumetric data is 
partitioned into non-overlapping patches ($16^3$) and projected into 
a semantic sequence via 3D patch embedding.
(B) Multi-modal Embedding: This module encodes auxiliary conditions to 
guide the generation. The target porosity $\phi$ is normalized 
via Z-score and mapped to the embedding dimension using a 
Multi-Layer Perceptron (MLP). Simultaneously, time steps $t$ are 
encoded using sinusoidal positional embeddings. These feature vectors 
are broadcasted and fused with the spatial patch embeddings 
(Feature Fusion) to condition the reverse denoising process
}
\end{flushleft}
\end{figure}

\subsubsection{Physical Representation and Forward Diffusion Process}

Distinct from conventional machine learning paradigms that treat 
rock voxel grids as pseudo-grayscale images, 
this study strictly adheres to the binary physical nature of porous media. 
We address a fundamental modeling mismatch in existing methods: 
framing generation as pixel intensity regression. 
Such approaches force discrete phases (pore/matrix) into continuous grayscale values, 
artificially creating blurred boundaries that necessitate post-hoc thresholding. 
This "blur-then-truncate" strategy is prone to topological errors, 
where slight threshold deviations can cause the spurious closure 
of narrow percolation throats or the erasure of micropores. 
To mitigate this, our method discards the "grayscale compromise." 
By employing end-to-end binary modeling, 
the network predicts precise phase probabilities rather than ambiguous intensities, 
thereby avoiding the topological uncertainties inherent in threshold-based segmentation.

As illustrated in \textbf{Fig. \ref{fig:my_structur}(A)}, the input binary voxel field $x_0 \in \{0,1\}^{1 \times D \times H \times W}$ (where 0 denotes matrix and 1 denotes pore space) is linearly mapped to a normalized range of $[-1, 1]$. 
This zero-centering operation aligns the physical signal with the standard Gaussian noise distribution.

The forward diffusion process is modeled as a fixed Markov chain. 
To ensure a smoother decay of the signal-to-noise ratio (SNR) before structural degradation, we employ the Cosine Noise Schedule proposed by Nichol et al.\cite{nichol2021improved} (detailed in \ref{appendix:noise_schedule}). 
Leveraging the reparameterization trick, the noisy state $x_t$ at any arbitrary time step $t \sim \mathcal{U}[1, T]$ can be sampled directly as follows \citep{ho2020denoising}:

\begin{equation}
\label{eqn:forward_process}
    x_{t} = \sqrt{\bar{\alpha}_{t}} \cdot x_{\text{map}} + \sqrt{1-\bar{\alpha}_{t}} \cdot \epsilon, \quad \epsilon \sim \mathcal{N}(0, \mathbf{I})
\end{equation}

where $x_{\text{map}} = 2x_0 - 1$ and $\bar{\alpha}_t$ denotes the cumulative noise variance. 
This formulation enables the network to learn robust reconstruction across the full spectrum of noise levels.

\subsubsection{3D Patch Embedding Strategy}

Addressing the cubic computational complexity $O(N^3)$ inherent in 3D voxel data and aligning with the serialization requirement of Transformers, we introduce a \textbf{3D Patch Embedding} strategy \citep{dosovitskiy2020image}. 
This process compresses the sparse, high-dimensional physical space into a compact, low-dimensional semantic space.

As depicted in the bottom panel of \textbf{Fig. \ref{fig:my_structur}(A)}, the input noisy voxel volume $x_t \in \mathbb{R}^{1 \times 256^3}$ is partitioned into $N$ non-overlapping cubic patches. 
With a patch size of $p=16$, each flattened raw voxel block $\mathbf{v}_{\text{raw}}^{(i)} \in \mathbb{R}^{p^3}$ is mapped to the embedding dimension $C$ via a learnable linear projection:
\begin{equation}
\label{eqn:patch_embedding}
    \mathbf{y}^{(i)} = \mathbf{v}_{\text{raw}}^{(i)} \mathbf{W}_{\text{proj}} + \mathbf{b}_{\text{proj}}, \quad i = 1, \dots, N
\end{equation}
where $\mathbf{W}_{\text{proj}} \in \mathbb{R}^{4096 \times C}$ is the projection matrix. 
Through this transformation, the discrete $16^3$ binary voxel block is compressed into a continuous $C$-dimensional floating-point vector. 
Physically, this vector functions as a ``feature fingerprint'' of the local volume, implicitly encoding the internal pore topology and connectivity. 
Consequently, the entire voxel field is reconstructed as a sequence $Z_0 \in \mathbb{R}^{N \times C}$. 
This spatial compression ratio of $4096:1$ enables the processing of Representative Elementary Volume (REV) scale data on consumer-grade hardware. 
Detailed network hyperparameters are provided in \textbf{\ref{appendix:arch_details}}.

\subsubsection{Multi-modal Embedding}

Since the Transformer architecture is inherently permutation-invariant, we incorporate learnable positional encodings to endow the model with \textbf{spatial awareness} of the 3D rock microstructure. 
As illustrated in \textbf{Fig. \ref{fig:my_structur}(B)}, a learnable position embedding parameter $E_{\text{pos}} \in \mathbb{R}^{N \times C}$ is added element-wise to the patch embedding sequence $Z_0$:
\begin{equation}
    Z_{\text{in}} = Z_0 + E_{\text{pos}}
\end{equation}
To enable physically controllable generation via \textbf{Classifier-Free Guidance (CFG)} \citep{ho2022classifier}, the model requires explicit conditioning on the noise level and physical properties. 
We explicitly map two scalar inputs---the \textbf{time step $t$} and the \textbf{target porosity $\phi$}---into the latent space. 

Adopting the frequency embedding strategy from Peebles et al.\cite{peebles2023scalable}, we treat both continuous scalars with a unified approach. 
The time step $t$ and normalized porosity $\phi$ are first expanded into high-frequency feature vectors via sinusoidal functions, and subsequently projected into global condition vectors $c_t$ and $c_{\phi}$ via Multi-Layer Perceptrons (MLP). 
To facilitate CFG during inference, we apply probabilistic dropout to these conditions during training (detailed settings are provided in \textbf{\ref{appendix:cfg_training}}). 
Consequently, the voxel sequence $Z_{\text{in}}$ combined with the condition vectors $\{c_t, c_{\phi}\}$ form the multi-modal input for the subsequent blocks.

\vspace{2em} 
\noindent
\begin{minipage}{\textwidth}
\centering
\includegraphics[width=\textwidth, height=0.8\textheight, keepaspectratio]{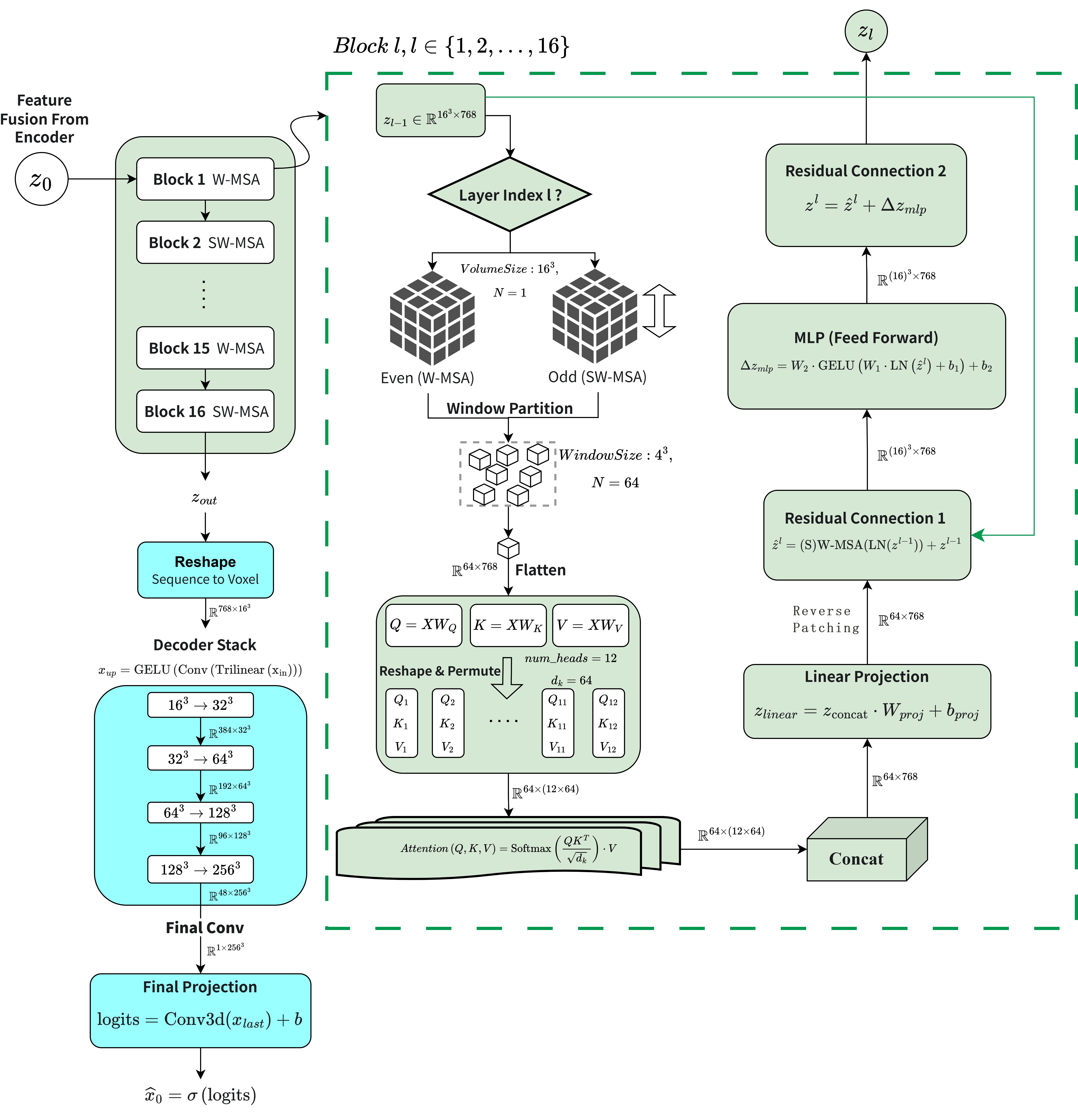}

\captionof{figure}{Architecture of the PoreDiT: Isotropic Swin-Transformer Encoder and Asymmetric Decoder}
\label{transformer} 

\begin{flushleft}
  {\footnotesize
The PoreDiT framework is composed of an isotropic Swin-Transformer encoder The framework comprises an isotropic encoder (left) composed of $L=16$ stacked Swin-Transformer blocks, maintaining a constant latent resolution of $16^3$ to preserve deep microscopic details. The right panel details the internal mechanism of the $l$-th block, which alternates between Window-based Multi-head Self-Attention (W-MSA) and Shifted Window (SW-MSA) to facilitate cross-window information interaction while maintaining linear computational complexity. The bottom panel depicts the lightweight asymmetric decoder, which progressively reconstructs the high-resolution voxel field ($256^3$) from the feature sequence $z_{out}$ via trilinear upsampling and convolutional refinement, finally outputting the phase probability map $\hat{x}_0$.
}
\end{flushleft}
\end{minipage}
\vspace{2em} 

\FloatBarrier

\subsection{PoreDiT Architecture}

\subsubsection{Isotropic Swin-Transformer Encoder}

As illustrated in Figure \ref{transformer}, our encoder eschews the ``compress-restore'' hourglass structure typical of U-Nets in favor of an isotropic deep architecture. The encoder comprises a stack of $L=16$ Swin-Transformer blocks, maintaining a constant spatial resolution of $16^3$ (4096 tokens) and a fixed feature dimension of $C=768$ throughout the network. This ``constant resolution'' strategy is critical for porous media, where hydraulic connectivity is predominantly governed by narrow pore throats; any form of downsampling (pooling) risks the irreversible loss of these high-frequency topological details.

To address computational bottlenecks when processing high-dimensional volumetric data, we adopt a window-based attention mechanism \citep{liu2021swin, liu2022video}. The computational flow for two consecutive blocks ($l$ and $l+1$) is formalized as:
\begin{equation}
\begin{aligned}
    & \hat{z}^l = \text{W-MSA}(\text{LN}(z^{l-1})) + z^{l-1}, \\
    & z^l = \text{MLP}(\text{LN}(\hat{z}^l)) + \hat{z}^l, \\
    & \hat{z}^{l+1} = \text{SW-MSA}(\text{LN}(z^l)) + z^l, \\
    & z^{l+1} = \text{MLP}(\text{LN}(\hat{z}^{l+1})) + \hat{z}^{l+1}
\end{aligned}
\label{eq:swin_block}
\end{equation}
where W-MSA (Window-based Multi-head Self-Attention) restricts computation within non-overlapping local 3D windows, reducing complexity from $O(N^2)$ to $O(N)$ (detailed mathematical formulations are provided in \ref{appendix:transformer}). To prevent the formation of ``information islands'' caused by isolated windows, we utilize SW-MSA (Shifted Window MSA) in alternating blocks. SW-MSA introduces cross-window connections by cyclically shifting the window partitioning along the $(D, H, W)$ axes. From the perspective of Percolation Theory, the stacking of SW-MSA layers facilitates the cross-domain diffusion of local information. This ensures that the effective receptive field in deeper layers expands to encompass the entire $256^3$ volume, allowing the model to simultaneously capture microscopic pore details and reconstruct complex, long-range global connectivity networks.

\subsubsection{Asymmetric Decoder \& Logit Projection}

To avoid the computational redundancy of symmetric decoders in traditional U-Nets, we adopt an asymmetric design inspired by \textbf{MAE \citep{he2022masked}}. As illustrated in Figure \ref{transformer}, our decoder is designed as a lightweight \textbf{Final Projection Block}. Unlike the deep Swin-Transformer encoder, this decoder contains no convolution or self-attention layers, focusing solely on mapping deep latent features back to the voxel space.

Specifically, the decoder first injects conditional information (e.g., porosity $\phi$) via \textbf{AdaLN (Adaptive Layer Normalization)}, followed by a \textbf{GELU} activation function to introduce non-linearity, and finally a linear layer for output. For the latent vector $z^L$ from the encoder, the decoding process is formalized as:
\begin{equation}
    \hat{v} = \text{Linear}(\text{GELU}(\text{AdaLN}(z^L, c)))
    \label{eq:decoder_block}
\end{equation}
where $c$ is the condition vector. The inclusion of \textbf{GELU (Gaussian Error Linear Unit)} is critical; it smooths the distribution in the latent space, enabling the linear projection layer to accurately reconstruct the sharp boundaries of binary pores (see \ref{appendix:decoder} for mathematical details). This design ensures parameter efficiency while maintaining high-fidelity reconstruction through non-linear activation.

\subsection{Optimization Strategy and Physics-Constrained Loss}
\label{sec:loss}

To overcome the non-differentiability of discrete binary voxels, we compute all losses directly on the continuous probability field $P(x) \in [0, 1]$ activated by the Sigmoid function. This relaxation strategy ensures that gradients from physical descriptors can backpropagate end-to-end, guiding the network optimization.

The total objective function consists of a reconstruction term and a physics-constrained term: $\mathcal{L}_{\text{total}} = \mathcal{L}_{\text{rec}} + \lambda_{\text{phy}} \mathcal{L}_{\text{phy}}$. The reconstruction loss $\mathcal{L}_{\text{rec}}$ combines Binary Cross-Entropy (BCE) and Dice loss to address both voxel-level accuracy and morphological overlap (see \ref{appendix:loss} for details).

For the physics-constrained loss $\mathcal{L}_{\text{phy}}$, relying solely on first-order statistics (such as porosity $\phi$) is often insufficient to capture complex pore-throat topological differences in generalization tasks. To enhance the model's perception of microstructure, we introduce the \textbf{Two-Point Correlation Function ($S_2$)} as a supplementary constraint \citep{torquato2002random}. We inject both $\phi$ and $S_2$ as equivalent physical conditions into the loss function:
\begin{equation}
    \mathcal{L}_{\text{phy}} = \| \hat{\phi} - \phi_{\text{gt}} \|^2 + \lambda_{s2} \sum_{r \in \mathcal{R}} \| \hat{S}_2(r) - S_2(r)_{\text{gt}} \|^2
    \label{eq:physics_loss}
\end{equation}
where $\mathcal{R}$ is a set of predefined spatial lag distances, and $\lambda_{s2}$ balances the gradient contributions. By jointly optimizing these statistics, the model is encouraged to recover not only the correct porosity but also topological consistency across diverse geological structures.

Based on the formulated objective, the complete training procedure of PoreDiT is summarized in summarized in \textbf{Figure \ref{alg:training}} (Algorithm 1). At each step, the model learns to directly recover the clean binary rock structure $x_0$ from the corrupted input.

\begin{algorithm}[t]
\caption{Training of PoreDiT}
\label{alg:training}
\begin{algorithmic}[1]
      \State \textbf{Input:} Dataset of binary rock volumes $\mathcal{D}$, Diffusion steps $T$, Noise schedule $\alpha_t, \sigma_t$
  \State \textbf{Output:} Optimized model parameters $\theta$
  \newline
  
  \While{not converged}
    \State Sample a real rock volume $x_0 \sim \mathcal{D}$;
    \State Sample time step $t \sim \text{Uniform}(1, \dots, T)$;
    \State Sample noise $\epsilon \sim \mathcal{N}(0, \mathbf{I})$;
    \State Corrupt input: $x_t = \alpha_t x_0 + \sigma_t \epsilon$;
    \State Predict original volume: $\hat{x}_0 = f_\theta(x_t, t)$; \textit{// Direct $x_0$ prediction}
    \State Calculate loss: $\mathcal{L} = \| x_0 - \hat{x}_0 \|^2$; 
    \State Update $\theta$ using gradient descent $\nabla_\theta \mathcal{L}$;
  \EndWhile
\end{algorithmic}
\end{algorithm}

\subsection{Physics-Constrained Inference and Large-scale Extrapolation}
\label{sec:inference}

\begin{figure}[pos=htbp]
\centering
\includegraphics[width=0.8\textwidth]{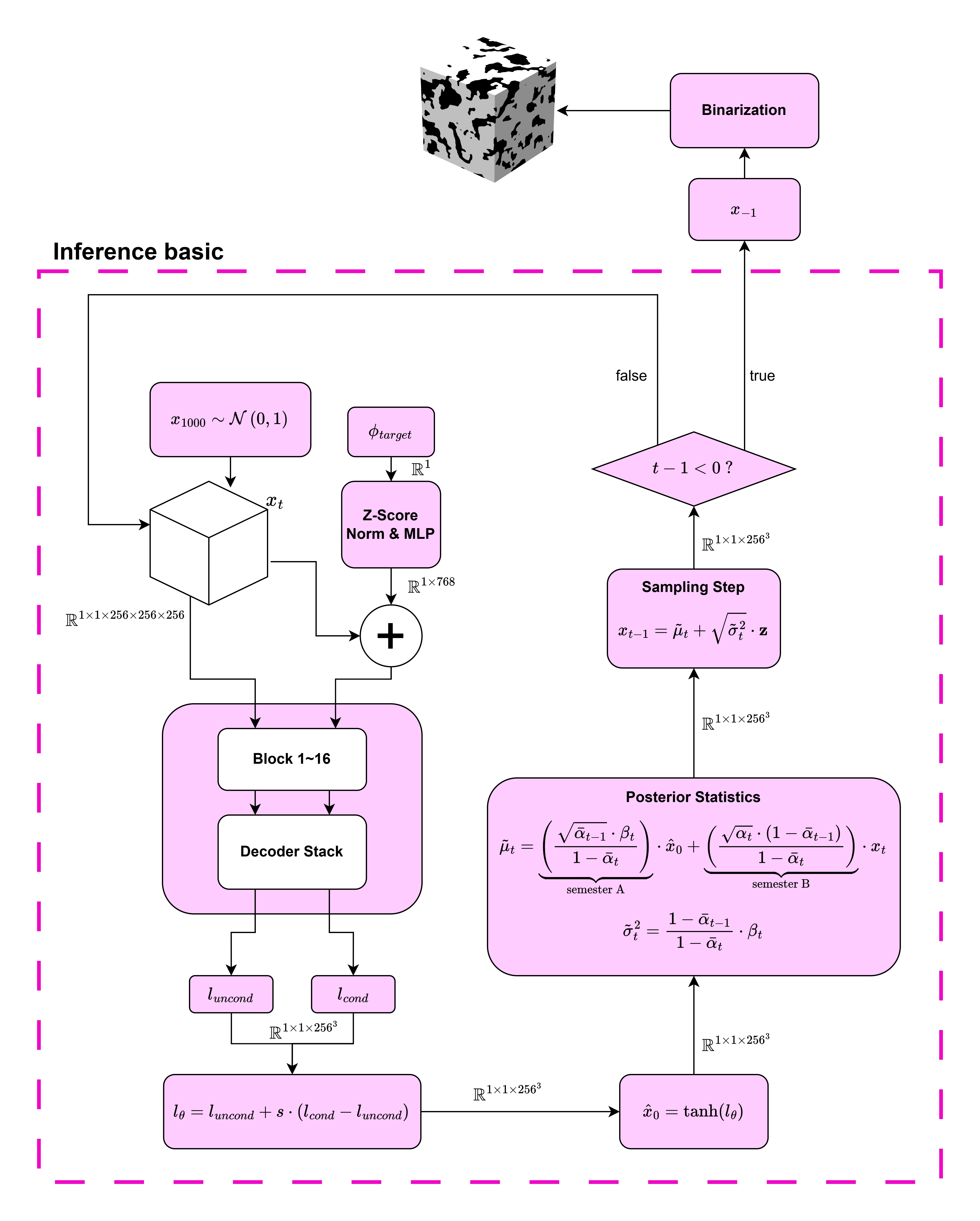} 
\caption{Schematic diagram of the reverse denoising inference process of the PoreDiT model.}
\label{inference} 
\begin{flushleft}
    {\footnotesize
The inference initiates with isotropic Gaussian noise $x_{1000}$, 
conditioned by the target porosity $\phi_{target}$. 
The parallel branches $l_{uncond}$ and $l_{cond}$ represent the 
unconditional and conditional logits, respectively, 
which are combined via the formula for $l_\theta$ to implement 
Classifier-Free Guidance (CFG) with scale $s$. 
The term $\hat{x}_0$ denotes the predicted clean state mapped to the 
$[-1, 1]$ interval. The block on the right illustrates the 
single-step sampling mechanism based on the posterior mean 
$\tilde{\mu}_t$ and variance $\tilde{\sigma}_t^2$. 
This iterative process repeats until $t=0$, 
followed by a final binarization step to recover the discrete 
pore-matrix structure.
    }
\end{flushleft}
\end{figure}

\subsubsection{Physics-Guided Sampling via $x_0$-Prediction}
\label{CFG}
As illustrated in Figure \ref{inference}, inference initiates by sampling Gaussian noise $x_T \sim \mathcal{N}(0, I)$. Unlike common diffusion models that predict noise $\epsilon$, our VoxelDiT is optimized to directly predict the original data $x_0$. At each denoising step $t$, the model estimates the clean volume $\hat{x}_0$ from the current state $x_t$ and physical condition $c$ (e.g., porosity $\phi$) by mapping the output logits to the $[-1, 1]$ interval via a Tanh activation. The mean of the previous state $\mu_{t-1}$ is then computed using the Bayesian posterior formulation (see \ref{appendix:inference_details} for details).

To enable controllable generation, we incorporate a \textbf{Classifier-Free Guidance (CFG)} mechanism during training by randomly dropping the condition $c$. During inference, the final logits are typically a linear combination of conditional and unconditional terms. However, for experiments on single lithologies (e.g., Bentheimer sandstone), we observed that the model fits the statistical distribution exceptionally well. Consequently, we disable strong guidance (setting the guidance scale $w=0$) in this phase, relying solely on the conditional branch for high-quality reconstruction. The CFG module is activated in subsequent generalization experiments to address distribution shifts(\ref{ketton}).

\begin{figure}[pos=htbp]
\centering
\includegraphics[width=0.8\textwidth]{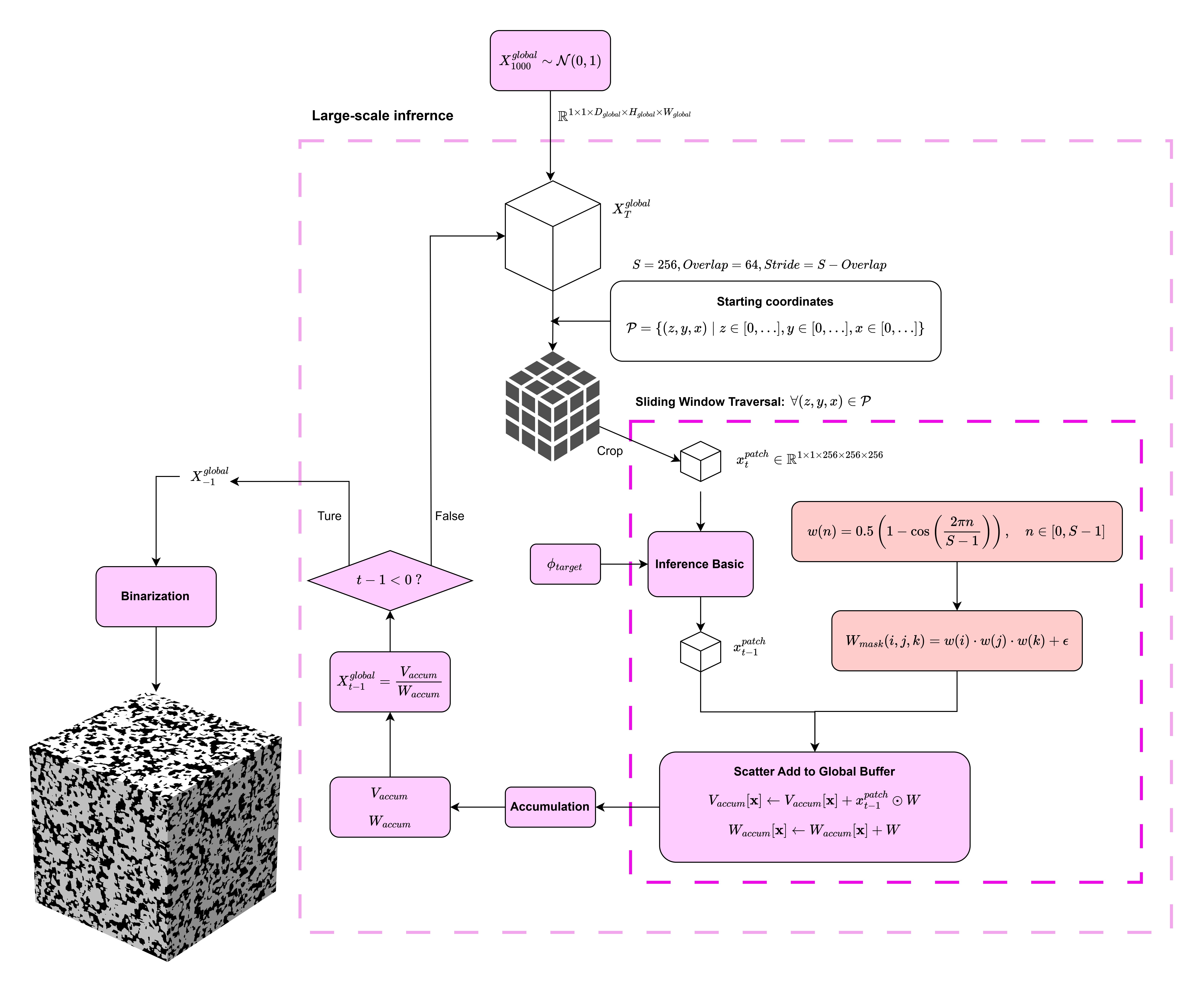} 
\caption{Schematic diagram of the reverse denoising inference process of the PoreDiT model.}
\label{largeinference} 
\begin{flushleft}
    {\footnotesize
The process begins with a global Gaussian noise initialization $X_{1000}^{global}$. A sliding window of size $S=256$ traverses the volume with a defined overlap $O=64$ (Stride $= S-O$). The coordinate set $\mathcal{P}$ defines the trajectory of these windows. Inside the dashed pink box, each local patch undergoes the standard denoising inference (refer to Fig \ref{inference}). To prevent boundary artifacts, a 3D Hann window function $w(n)$ is applied to weight the outputs. The final global volume at the next timestep $X_{t-1}^{global}$ is reconstructed by normalizing the accumulated values ($V_{accum}$) by the accumulated weights ($W_{accum}$). This divide-and-conquer approach enables gigavoxel-scale generation on limited memory hardware.
    }
\end{flushleft}
\end{figure}

\subsubsection{Large-scale Extrapolation with Global Coherent Noise}
To generate large-scale digital cores (e.g., $1024^3$) exceeding consumer-grade GPU memory limits, we employ a sliding-window (tiled) sampling strategy. As shown in Figure \ref{largeinference}, the large volume is partitioned into overlapping patches, which are denoised independently and fused via weighted averaging.

However, conventional tiled diffusion leads to severe \textbf{Porosity Drift}. We identify that this stems from sampling Gaussian noise independently for each patch: averaging independent noise in overlap regions violates the ``unit variance'' assumption of the Gaussian distribution, causing variance collapse (see mathematical proof in \ref{appendix:inference_details}). To resolve this and maintain \textbf{Statistical Stationarity} across the entire volume, we propose a \textbf{Global Coherent Noise} strategy. Specifically, a global noise field $z^{\text{global}}$ is initialized once at $t=T$, and slices of this field are used to drive local generation steps. This simple yet effective correction mathematically guarantees variance conservation, eliminating stitching artifacts and ensuring consistent porosity distribution at large scales.

The detailed inference pipeline is outlined in \textbf{Figure \ref{algorithm:inference}}, which describes the iterative reconstruction from pure Gaussian noise to the final binarized structure $x_0$.

\begin{algorithm}[t]
\caption{Inference via Direct $x_0$ Prediction}
\label{algorithm:inference}
\begin{algorithmic}[1]
    \State \textbf{Input:} Model $f_\theta$, Total steps $T$, Gaussian noise $x_T \sim \mathcal{N}(0, \mathbf{I})$
  \State \textbf{Output:} Generated digital rock $x_0$
  \newline

  \For{$t = T, T-1, \dots, 1$}
    \State Predict clean image: $\hat{x}_0 = f_\theta(x_t, t)$;
    \State Estimate previous noise: $\hat{\epsilon}_t = (x_t - \alpha_t \hat{x}_0) / \sigma_t$;
    \State Compute previous state mean: $\mu_{t-1} = \alpha_{t-1} \hat{x}_0 + \sigma_{t-1} \hat{\epsilon}_t$;
    \State Sample noise $z \sim \mathcal{N}(0, \mathbf{I})$ if $t > 1$ else $z=0$;
    \State Update state: $x_{t-1} = \mu_{t-1} + \eta \sigma_{t-1} z$;
  \EndFor
  \State Binarize $x_0$ using Otsu's thresholding;
  \State \textbf{Return} $x_0$;
\end{algorithmic}
\end{algorithm}

\section{Results}
\subsection{Implementation Details}
This study utilizes the Bentheimer sandstone dataset (Project DRP-317) available on the Digital Rocks Portal as the primary data source \citep{neumann2020}. Bentheimer sandstone, a Lower Cretaceous shallow-marine deposit characterized by high quartz content and uniform texture, serves as a widely used benchmark for homogeneous sandstone analysis. The dataset consists of high-quality binary voxel images with dimensions of $1000 \times 1000 \times 1000$ voxels. To construct the training dataset, the original volume was partitioned into 216 sub-volumes, each with a size of $256 \times 256 \times 256$ voxels. This preprocessing stage incorporated overlapping cropping strategies and data augmentation techniques, such as flipping.

All experiments in this study were conducted on a Linux workstation 
equipped with two NVIDIA GeForce RTX 4090 GPU (24 GB VRAM). 
For the training phase, we employed a Data Parallelism strategy 
implemented via the PyTorch framework, utilizing the Hugging Face 
Accelerate library for distributed training management.To achieve 
efficient training on consumer-grade hardware, we adopted 
mixed-precision training (fp16) and the 8-bit AdamW optimizer from 
the bitsandbytes library. The training input consisted of voxel 
samples with a resolution of $256^3$. The per-GPU batch size was set 
to 1 (effective total batch size of 2), controlling the memory usage 
per card to approximately 18.3 GB. The learning rate was set to 
$3 \times 10^{-5}$, and the number of diffusion timesteps was set 
to 1000.

Regarding the loss function configuration, the total objective 
is a weighted sum of Binary Cross-Entropy (BCE) loss, Dice loss, 
Gradient loss (weight $\lambda_{grad}=0.008$), and Condition-guidance 
loss (weight $\lambda_{cond}=1.5$). To ensure training stability, 
a linear warm-up strategy was applied to the conditional loss during 
the first three epochs. The model was trained from scratch for 300 
epochs, with an average duration of approximately 2 minutes and 29 
seconds per epoch.

During the inference (sampling) phase, 
FP32 precision was used for $256^3$ and $512^3$ resolution samples 
to ensure generation quality. However, for the large-scale generation 
of $1024^3$ volumes, the sampling precision was adjusted to FP16 to 
accommodate VRAM constraints.

\subsection{ Validation on Standard Benchmark ($256^3$)}
To ensure the statistical representativeness and fairness of the 
validation, we first derived the empirical distribution of porosity 
from the original core samples. Based on this, we sampled a set of 
target porosity values as conditional inputs to generate 100 
representative synthetic samples. We established a multi-dimensional 
validation framework to comprehensively evaluate the performance of 
the proposed method and the authenticity of the generated 
microstructures. This framework encompasses four core aspects:
\begin{itemize}
\item \textbf{Visual Inspection:} Intuitively judging the morphological 
similarity between the generated pore structures and real rock samples.
\item \textbf{Morphological \& Statistical Analysis:} Quantifying 
geometric consistency by calculating the Two\-point Correlation 
Function ($S_2$) and Minkowski Functionals (MFs).
\item \textbf{Physical Property Evaluation:} Examining whether fluid 
transport properties, such as absolute permeability, 
conform to geological laws.
\item \textbf{Generative Quality and Diversity:} Verifying the model's 
generalization capability to create realistic "novel" samples rather 
than merely memorizing training data.
\end{itemize}

\subsubsection{ Visual Inspection}
\begin{figure}[pos=htbp]
    \centering
    
    \begin{minipage}{0.2\linewidth}
        \centering \textbf{Ground Truth (GT)}
    \end{minipage}
    \hspace{1.5cm} 
    \begin{minipage}{0.2\linewidth}
        \centering \textbf{Generated (Ours)}
    \end{minipage}
    
    \vspace{0.2cm} 

    \begin{subfigure}{0.25\linewidth}
        \centering
        \tikz\node[draw=red, dashed, line width=1pt, inner sep=0pt]{\includegraphics[width=\linewidth]{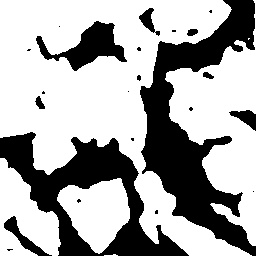}};
        \caption{} 
        \label{fig:a}
    \end{subfigure}
    \hspace{1.5cm} 
    \begin{subfigure}{0.25\linewidth}
        \centering
        \tikz\node[draw=red, dashed, line width=1pt, inner sep=0pt]{\includegraphics[width=\linewidth]{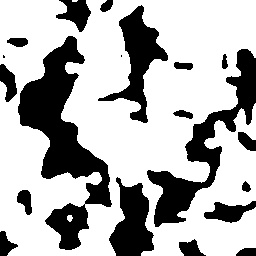}};
        \caption{} 
        \label{fig:b}
    \end{subfigure}

    \vspace{0.2cm}

    \begin{subfigure}{0.25\linewidth}
        \centering
        \tikz\node[draw=red, dashed, line width=1pt, inner sep=0pt]{\includegraphics[width=\linewidth]{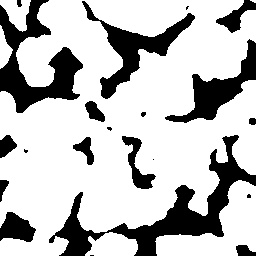}};
        \caption{} 
        \label{fig:c}
    \end{subfigure}
    \hspace{1.5cm}
    \begin{subfigure}{0.25\linewidth}
        \centering
        \tikz\node[draw=red, dashed, line width=1pt, inner sep=0pt]{\includegraphics[width=\linewidth]{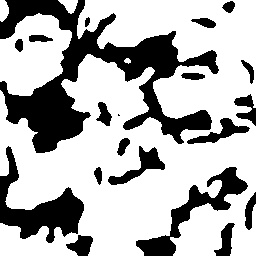}};
        \caption{} 
        \label{fig:d}
    \end{subfigure}

    \vspace{0.2cm} 

    \begin{subfigure}{0.25\linewidth}
        \centering
        \tikz\node[draw=red, dashed, line width=1pt, inner sep=0pt]{\includegraphics[width=\linewidth]{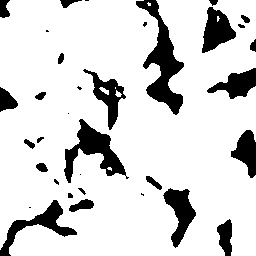}};
        \caption{} 
        \label{fig:e}
    \end{subfigure}
    \hspace{1.5cm}
    \begin{subfigure}{0.25\linewidth}
        \centering
        \tikz\node[draw=red, dashed, line width=1pt, inner sep=0pt]{\includegraphics[width=\linewidth]{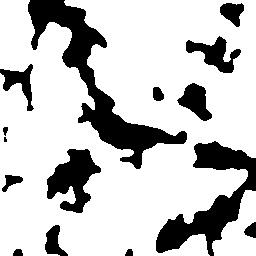}};
        \caption{} 
        \label{fig:f}
    \end{subfigure}

    \caption{Visual comparison of microscopic structures.
    The left column displays 2D cross-sections ($256 \times 256$ pixels) extracted from the ground truth volumes, and the right column shows the corresponding synthetic slices generated by our model.
    The rows correspond to samples with high (a, b), average (c, d), and low (e, f) porosity levels, respectively.}
    \label{fig:visual_comparison}
\end{figure}

Visual inspection serves as the primary assessment of the fidelity 
of generated microstructures. Figures \ref{fig:visual_comparison}(a), \ref{fig:visual_comparison}(b), \ref{fig:visual_comparison}(c), \ref{fig:visual_comparison}(d), \ref{fig:visual_comparison}(e), and \ref{fig:visual_comparison}(f) present a comparison of 
typical 2D slices extracted from the 3D training volumes and generated 
volumes. Although slices display only partial information, they 
clearly reflect the complex pore topology. As shown, the samples 
generated by our method (right column) are visually highly consistent 
with the real rock samples (left column). The model accurately 
reproduces the spatial distribution patterns of pores (black regions) 
and the solid matrix (white regions), successfully capturing local 
heterogeneity under different target porosities without exhibiting 
obvious artifacts or mode collapse.

\begin{figure}[pos=htbp]
    \centering
    
    \begin{subfigure}{0.4\linewidth}
        \centering
        \includegraphics[width=\linewidth]{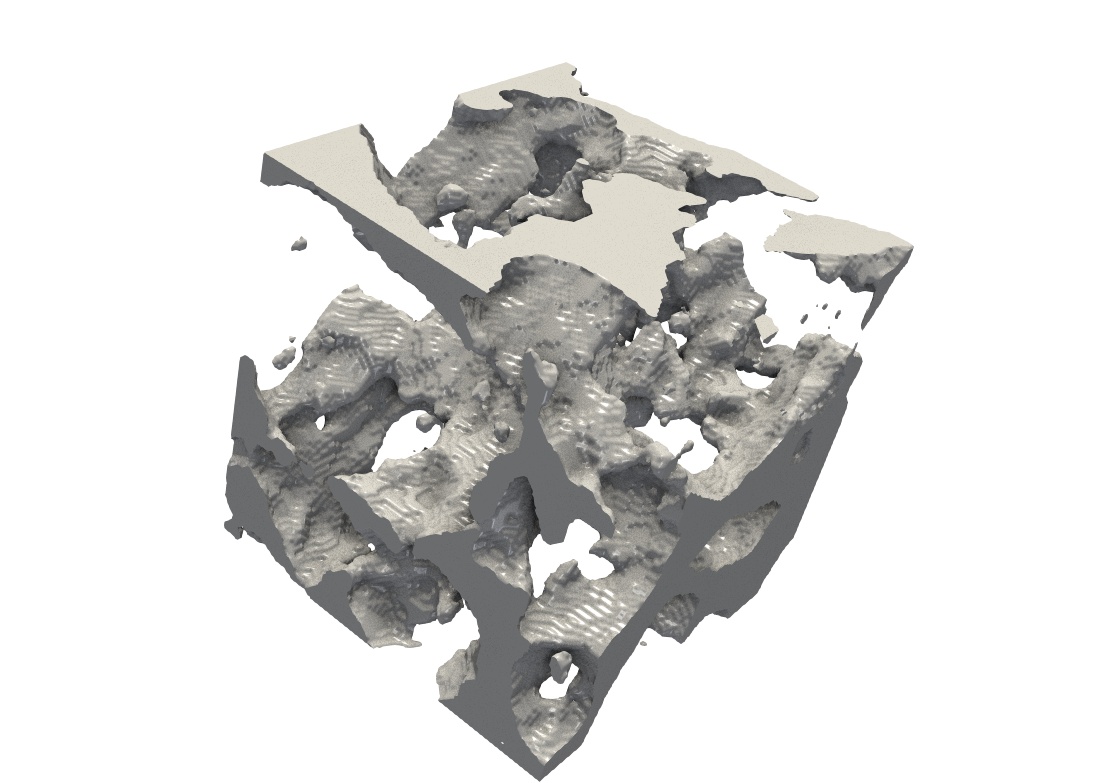} 
        \caption{Ground Truth (GT)}
        \label{fig:3d_gt}
    \end{subfigure}

    \begin{subfigure}{0.4\linewidth}
        \centering
        \includegraphics[width=\linewidth]{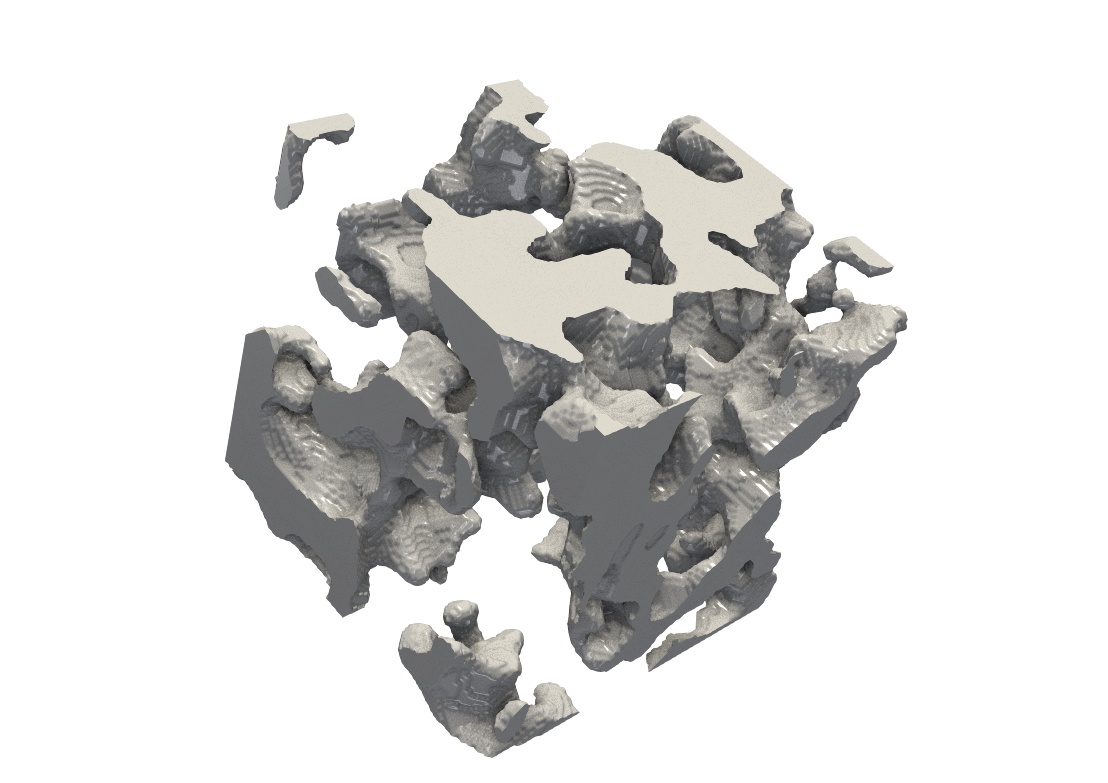} 
        \caption{Generated (Ours)}
        \label{fig:3d_gen}
    \end{subfigure}
    
    \caption{Comparison of 3D pore network structures. (a) Isosurface rendering of a real sandstone sub-volume ($128^3$ voxels) randomly extracted from the training set. (b) A synthetic sub-volume ($128^3$ voxels) generated by the PoreDiT model. Gray solid regions represent the pore space, illustrating the effective reconstruction of topological connectivity.}
    \label{fig:3d_render_comparison}
\end{figure}

Given that 2D slices cannot fully characterize the volumetric 
connectivity and anisotropy of the pore space, we further performed 
3D volume rendering visualization. Figures \ref{fig:3d_render_comparison}(a) and \ref{fig:3d_render_comparison}(b) compare the isosurface 
renderings of real rock sub-volumes from the training set (left) and 
synthetic sub-volumes generated by the model (right). Observations 
indicate that the generative model successfully reproduces the complex 
3D topological structure of the rock. Notably, the generated pore 
networks demonstrate extension trends and connectivity in 3D space 
that are highly similar to real samples, with no isolated noise points 
or layer-wise discontinuities. This strongly evidences that the model 
has captured long-range spatial correlations between voxels rather 
than merely learning 2D texture distributions.

\subsubsection{Morphological \& Statistical Analysis}
\label{sec:morphological_analysis}

\begin{figure}[pos=htbp]
\centering
\includegraphics[width=0.6\textwidth]{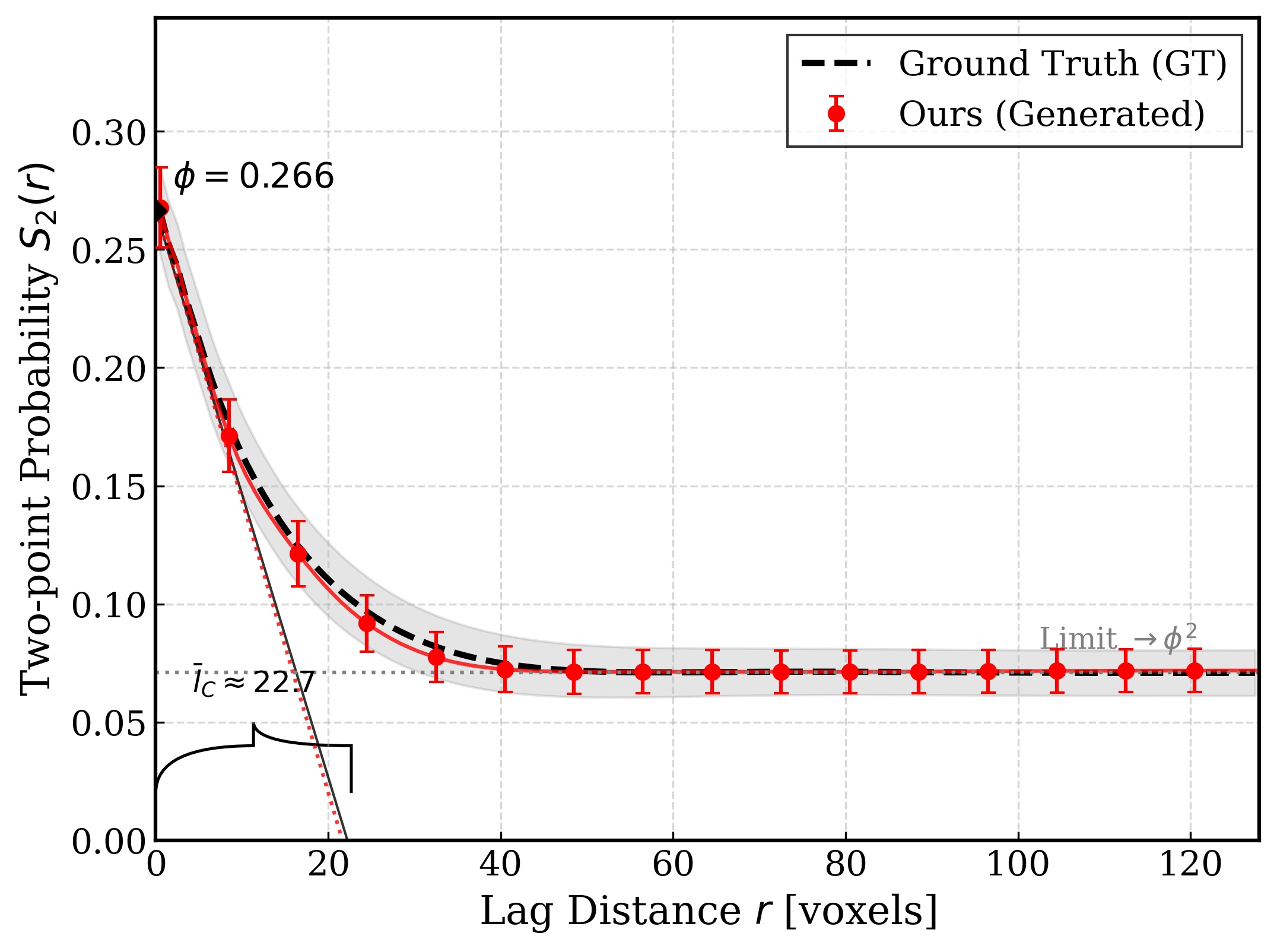} 
\caption{
Comparison of the two-point correlation function $S_2(r)$ between the ground truth and generated samples.}
\label{fig:s2_plot} 
\begin{flushleft}
    {\footnotesize
The red solid points represent the mean values of the generated samples 
(Ours), with error bars indicating the standard deviation. 
The black dashed line denotes the average $S_2(r)$ of the ground truth 
(GT) training samples, while the grey shaded region represents the 
corresponding standard deviation range of the GT data. Key statistical 
features are annotated as follows: the Y-axis intercept corresponds to 
the porosity ($\phi \approx 0.266$); the slope of the tangent at the 
origin (thin straight lines) reflects the specific surface area; the 
intersection of the tangent with the X-axis indicates the 
characteristic correlation length ($\bar{l}_c \approx 22.7$ voxels); 
and the horizontal asymptote at large lag distances represents the 
theoretical convergence limit $\phi^2$.
    }
\end{flushleft}
\end{figure}
To quantitatively assess the geometric and topological fidelity of the generated microstructures, we benchmarked the generated samples against the Ground Truth (GT) using the Two-point Correlation Function ($S_2$) and Minkowski Functionals, following the standards established by Mosser et al.\cite{mosser2017reconstruction}. Detailed calculation methodologies are provided in \ref{appendix_s2} and \ref{appendix_minkowski}.

\paragraph{Two-point Correlation Function ($S_2$)}
Figure~\ref{fig:s2_plot} compares the $S_2(r)$ curves of the generated volumes with the real Bentheimer sandstone. The high degree of overlap verifies the physical validity of the model across scales: (1) Near the origin ($r \to 0$), the matching slopes indicate that the generated pores possess specific surface area and interfacial roughness comparable to the real rock; (2) At large distances ($r \to \infty$), the generated curve converges stably to $\phi^2$ (where $\phi$ is porosity) without significant oscillation. This confirms that \textbf{PoreDiT} captures not only local pore morphology but also the stationarity and long-range correlations inherent in the rock, effectively avoiding mode collapse or periodic artifacts common in generative models.

\begin{figure}[pos=htbp]
    \centering
    \includegraphics[width=\linewidth]{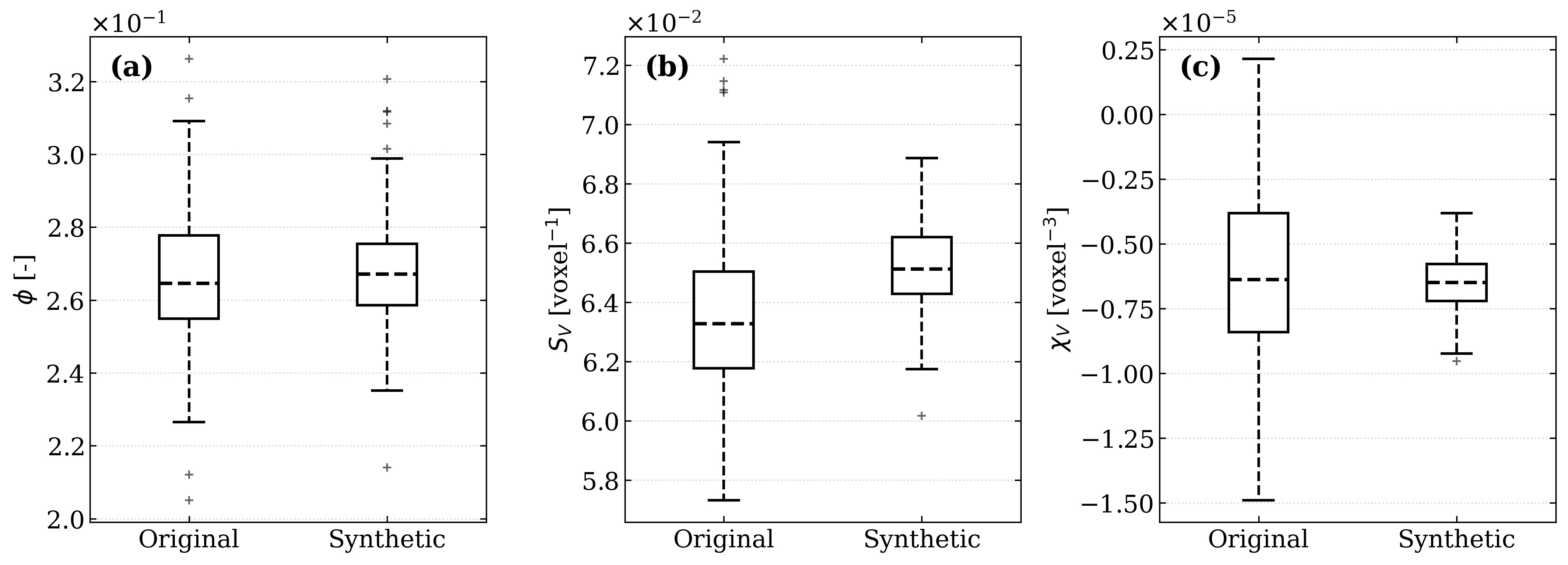}
    \caption{Comparison of Minkowski Functionals (MFs) distributions between original training samples and synthetic samples. The panels display the statistical distributions of (a) Porosity $\phi$ (corresponding to $M_0$), (b) Specific Surface Area $S_V$ (corresponding to $M_1$), and (c) Euler Characteristic Density $\chi_V$ (corresponding to $M_2$). The boxes represent the interquartile range (IQR), the central dashed lines indicate the median, whiskers extend to the full data range (excluding outliers), and '+' symbols denote outliers.}
    \label{fig:minkowski_boxplots}
\end{figure}

\paragraph{Minkowski Functionals}
Figures~\ref{fig:minkowski_boxplots}(a), \ref{fig:minkowski_boxplots}(b), and \ref{fig:minkowski_boxplots}(c) present boxplots comparing three key topological descriptors: Porosity ($V_0$), Specific Surface Area ($V_1$), and Euler Characteristic ($V_2$). The statistical distributions (medians and interquartile ranges) of the generated samples align closely with the Ground Truth. Notably, for the Euler Characteristic ($V_2$), which characterizes connectivity, the range of the generated samples fully overlaps with that of the real samples after removing non-physical noise clusters (see \ref{appendix_minkowski}). This strongly evidences that the generated pore networks are topologically connected and natural, rather than a mere accumulation of visually similar but isolated pores.

\subsubsection{Permeability Evaluation via Lattice Boltzmann Method}
\label{sec:permeability}

\begin{figure}[pos=htbp]
    \centering
    \includegraphics[width=\linewidth]{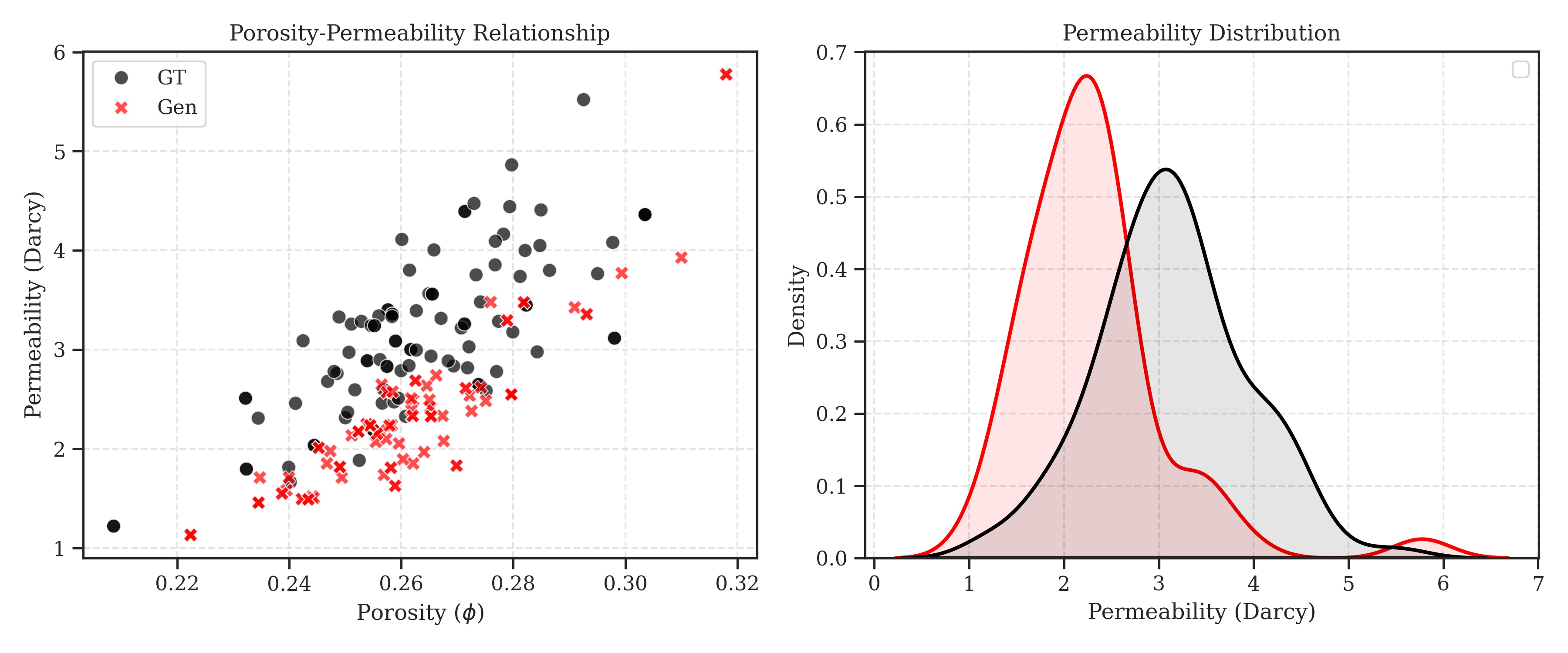}
    \caption{Permeability comparison via LBM simulations. The left panel shows the porosity-permeability relationship, indicating that the generated samples fall within the valid physical range of the ground truth. The right panel displays the probability density of permeability; despite a slight leftward shift in the peak ($\sim 0.8$ Darcy offset), the generated distribution (red) shows high structural similarity to the ground truth (black) and lies within its statistical support.}
    \label{fig:bentheimer_perm}
\end{figure}

To evaluate the fluid transport capacity of the generated pore networks, we employed the Lattice Boltzmann Method (LBM) to directly simulate single-phase fluid flow at the pore scale. Detailed numerical settings, boundary conditions, and physical parameter derivations are provided in \ref{appendix_lbm} \citep{Succi2001}.

Figure~\ref{fig:bentheimer_perm} compares the absolute permeability distributions of the generated samples against the Ground Truth along three axial directions. Statistical results indicate that the samples generated by \textbf{PoreDiT} successfully cover the permeability range of real Bentheimer sandstone (approximately 1000--3000 mD), and the overall distribution shape (e.g., skewness and spread) exhibits good consistency with the real data.

It is noted that the peak of the permeability distribution for the generated samples is slightly lower than that of the Ground Truth (a leftward shift about 0.8 Darcy). This suggests that the generated pore throats may be marginally more tortuous or constricted than those in the natural rock. Despite this minor statistical deviation, the generated samples maintain transport properties of the same order of magnitude, rendering them valid for subsequent hydrodynamic simulations. A detailed analysis of the potential structural reasons for this slight underestimation is presented in Section~\ref{limitations}.

\subsubsection{Generative Quality \& Diversity}
While previous experiments confirmed that the generated microstructures align with real rocks in terms of statistical features, topological connectivity, and physical properties, it is equally critical to verify the model's generalization capability. Specifically, we must rule out mere memorization or overfitting of the training data.

\begin{figure}[pos=htbp]
  \centering 
  \includegraphics[width=0.6\textwidth]{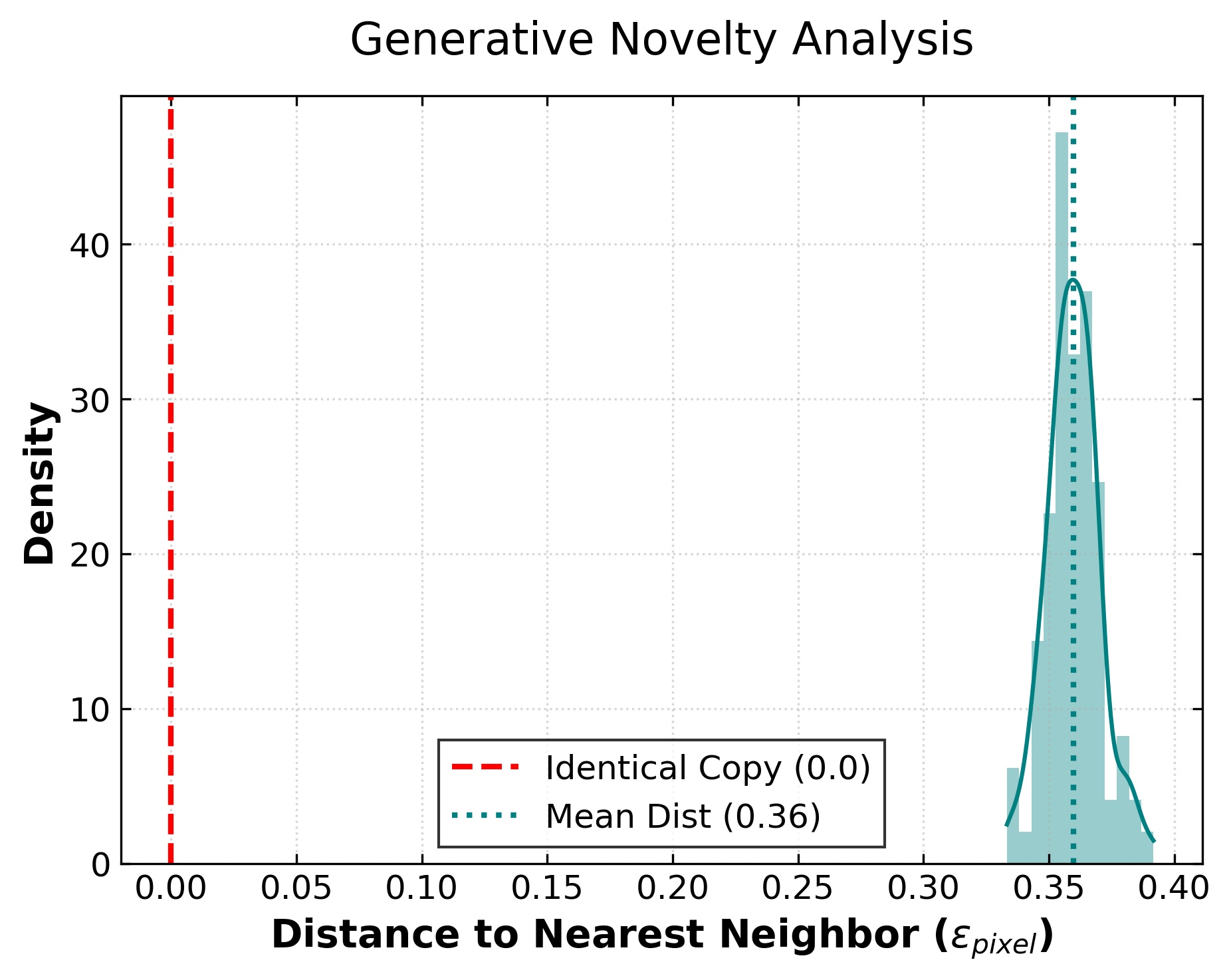} 
  \caption{\textbf{Generative Novelty Analysis.} 
  Histogram of the Distance to Nearest Neighbor ($D_{min}$) for generated samples. The X-axis represents the pixel-wise distance ($\epsilon_{pixel}$) to the nearest neighbor in the training set. The red dashed line ($x=0$) indicates the theoretical position for "exact data copying." The solid teal line represents the Kernel Density Estimation (KDE) curve, and the dotted line indicates the mean distance of the samples.}
  \label{fig:novelty_analysis}
\end{figure}

Following the benchmark framework established by \cite{lee2024microstructure}, we employed the Distance to Nearest Neighbor ($D_{min}$) as a quantitative metric to detect "data-copying." This metric is defined as the minimum pixel-wise difference between a generated sample $G_i$ and its most similar counterpart in the entire training dataset $\mathcal{R}$:
\begin{equation} D_{min}^{(i)} = \min_{R_j \in \mathcal{R}} | G_i - R_j | \end{equation}
Figure \ref{fig:novelty_analysis} presents the probability distribution histogram of $D_{min}$. Theoretically, "exact copying" corresponds to $D_{min} \to 0$. The experimental results show that the distribution centers primarily within the interval $[0.32, 0.40]$, with a mean of approximately 0.36. Crucially, a distinct gap exists between the lower bound of the distribution (minimum value $\ge 0.33$) and zero. This significant separation indicates that even the generated sample most similar to the training set exhibits substantial structural differences in the pixel space. This strongly evidences that the PoreDiT model does not merely splice or memorize data but has successfully learned the underlying distribution patterns of rock microstructures, possessing the capability to synthesize novel and diverse samples.

\begin{figure}[pos=htbp]
\centering
\includegraphics[width=\linewidth]{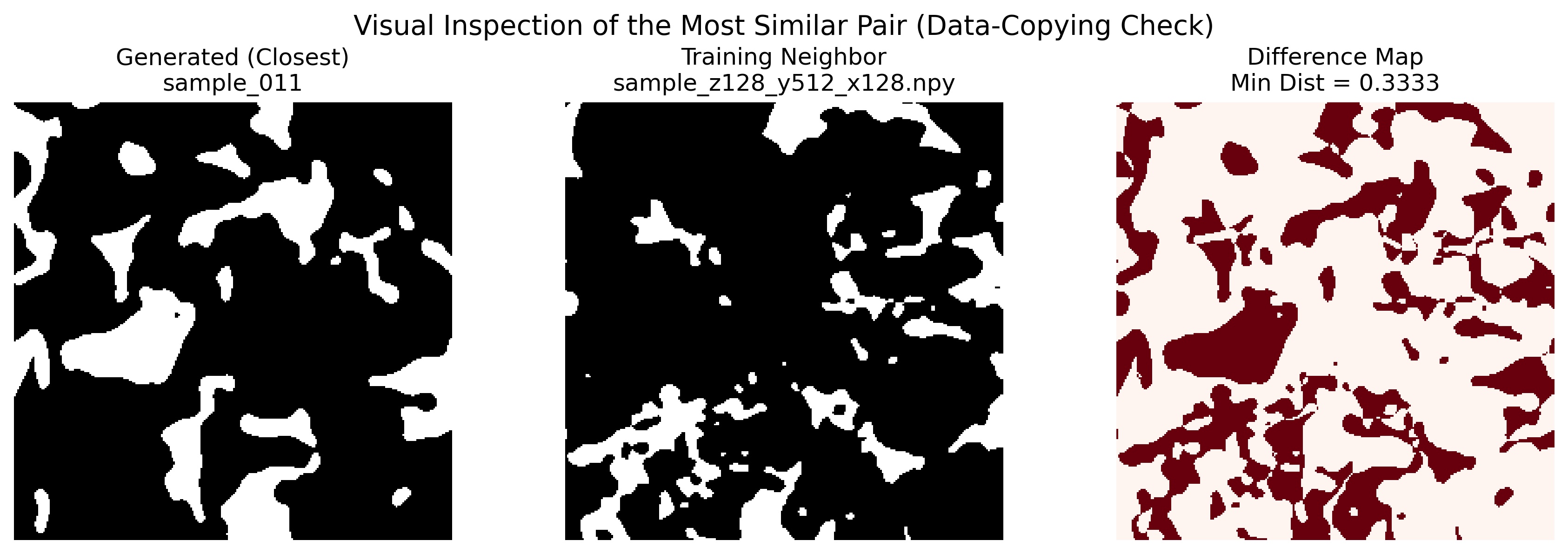}
\caption{\textbf{Visual inspection of the most similar pair (Data-Copying Check).} (Left) The generated sample with the minimum distance metric. (Middle) The nearest neighbor sample identified from the training set. (Right) The pixel-wise difference map between the two. High-intensity red regions indicate significant pixel residuals ($D_{min} \approx 0.33$), confirming structural distinctness.}
\label{fig:visual_copy_check}
\end{figure}

To intuitively validate this quantitative analysis, Figure \ref{fig:visual_copy_check} visualizes the pair of samples with the highest similarity in the entire dataset (i.e., the extreme case corresponding to $D_{min}=0.33$). While the generated sample (left) and its training neighbor (middle) exhibit striking consistency in macroscopic texture and pore size distribution—explaining why the algorithm identified them as nearest neighbors—careful inspection reveals no point-to-point correspondence in their microstructures. The difference map (right) further confirms this, where widespread high-intensity (red) regions reveal substantial pixel-level residuals. This visualization reinforces that PoreDiT does not "recite" training data at the pixel level. Instead, it captures the underlying morphological patterns (style) of the rock, generating novel samples with independent topological structures while maintaining geological realism.

\subsection{Ablation Study: Controllability Analysis}
\label{sec:ablation_study}

To investigate the necessity of explicit physical guidance, we performed an ablation study comparing the unconditional baseline with our porosity-conditioned model.

\begin{figure}[pos=htbp]
\centering
\includegraphics[width=0.6\linewidth]{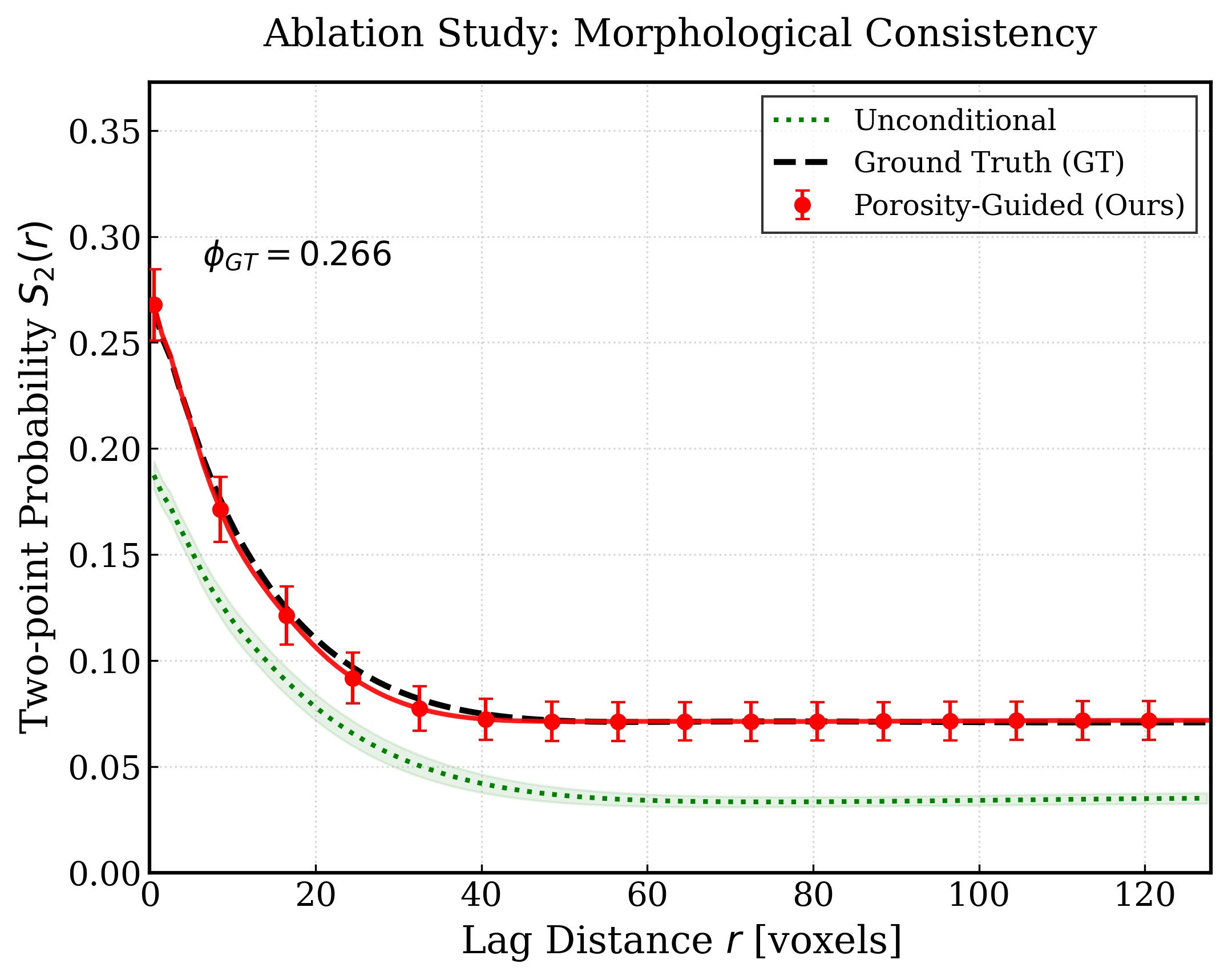}
\caption{\textbf{Ablation Study: Impact of physical guidance on morphological consistency ($S_2(r)$).} The green dotted line represents the Unconditional Baseline, the red solid points with error bars represent the Porosity-Guided model (Ours), and the black dashed line represents the Ground Truth (GT).}
\label{fig:ablation_s2}
\end{figure}

\begin{figure}[pos=htbp]
    \centering
    
    \begin{subfigure}{0.48\linewidth}
        \centering
        \includegraphics[width=\linewidth]{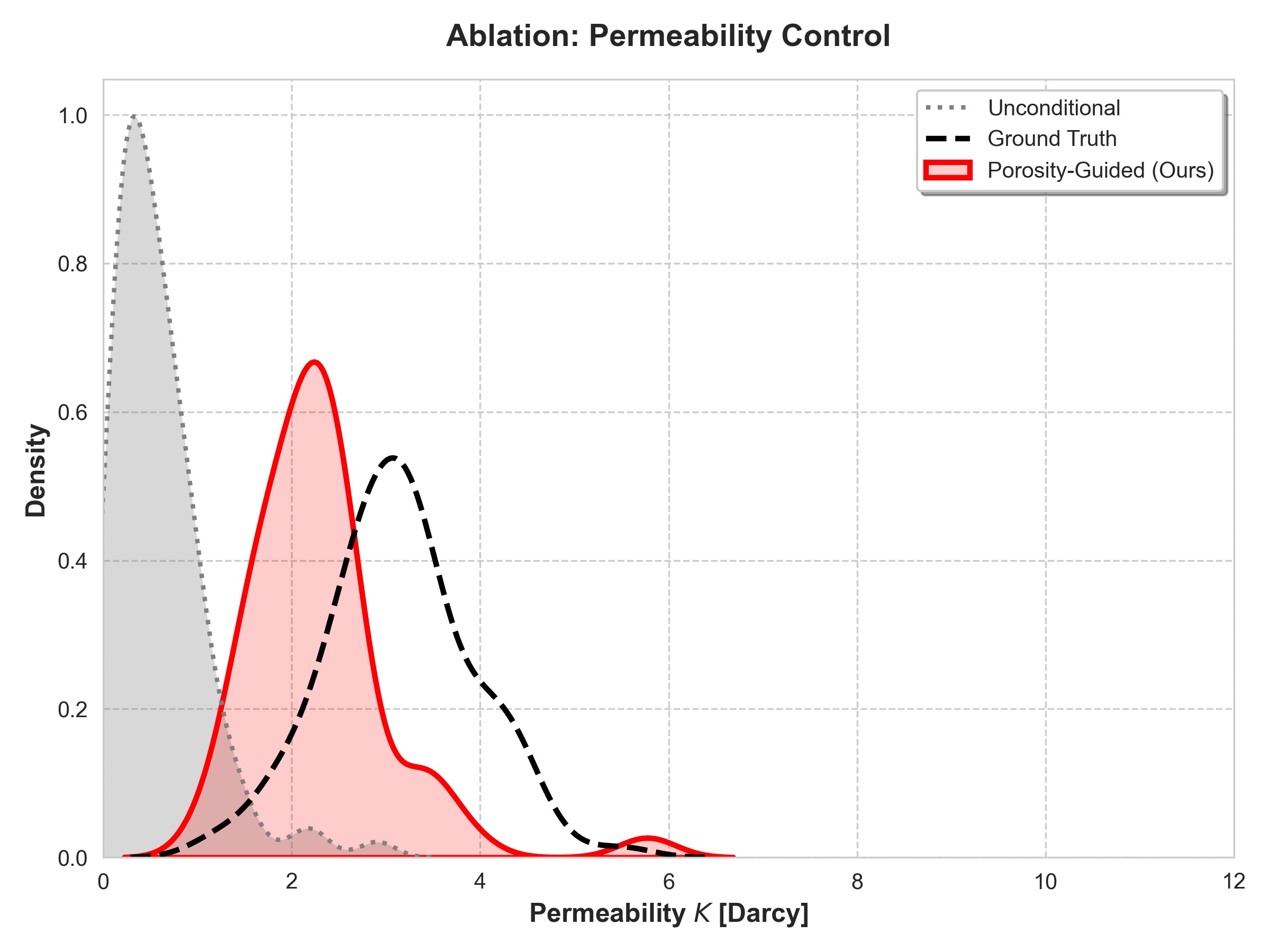} 
        \caption{}
        \label{fig:ablation_perm_pdf}
    \end{subfigure}
    \hfill 
    \begin{subfigure}{0.48\linewidth}
        \centering
        \includegraphics[width=\linewidth]{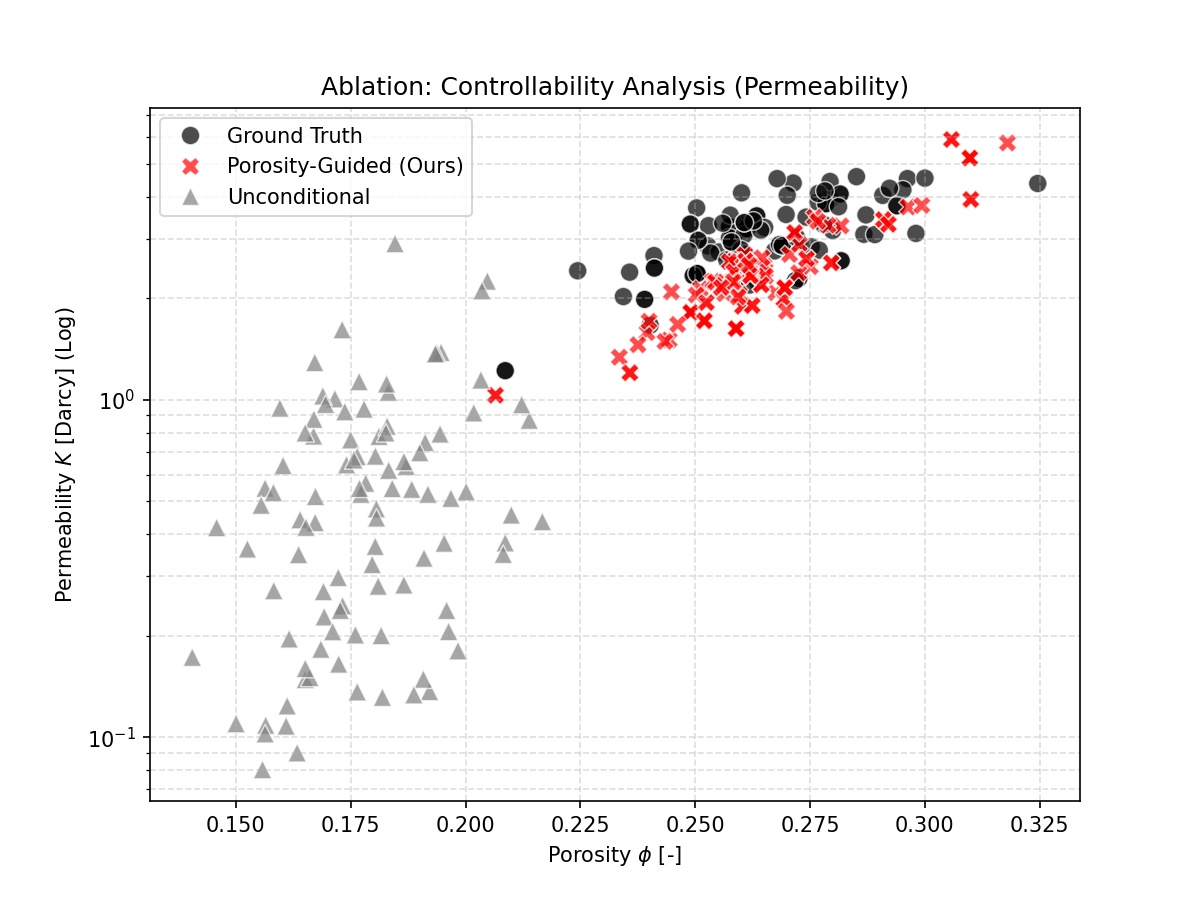} 
        \caption{} 
        \label{fig:ablation_perm_scatter}
    \end{subfigure}
    
    \caption{\textbf{Ablation Study: Controllability Analysis of Physical Properties.} 
    (a) Probability Density Function (PDF) of absolute permeability $K$. The grey shaded area, red curve, and black dashed line represent the Unconditional Baseline, Porosity-Guided model, and Ground Truth, respectively. 
    (b) Joint distribution of porosity $\phi$ and permeability $K$. Grey triangles, red crosses, and blue circles correspond to the Unconditional, Porosity-Guided, and Ground Truth samples, respectively.}
    \label{fig:ablation_perm}
\end{figure}

The quantitative results are presented in Figure~\ref{fig:ablation_s2} and Figures~\ref{fig:ablation_perm}(a) and \ref{fig:ablation_perm}(b). While the unconditional model is capable of generating visually plausible local textures, it suffers from significant ``Porosity Drift'' in terms of global statistical features. Specifically, the porosity distribution of the unconditional samples exhibits uncontrollable random fluctuations, failing to align precisely with the target mean of the real rock.

This deviation is further amplified in the physical property space. Using the same evaluation metrics established in Section~\ref{sec:morphological_analysis} and \ref{sec:permeability}, the unconditional samples show a divergent trend in the permeability-porosity scatter plot, deviating significantly from the inherent petrophysical relationship of the Bentheimer sandstone. In contrast, by incorporating the conditional guidance mechanism, the model not only precisely locks onto the target porosity (with error within $\pm 1\%$) but also yields $S_2$ curves and permeability distributions that tightly match the Ground Truth. This confirms that data-driven texture learning alone is insufficient for physical fidelity; injecting global physical constraints during the denoising process is indispensable for preventing statistical drift.

\subsection{Large-scale Reconstruction and Global Consistency Assessment (\texorpdfstring{$1024^3$}{$512^3$})}

Using the pre-trained PoreDiT model, we performed large-scale generation conditioned on the average porosity of the original dataset. The experiments were conducted on two NVIDIA RTX 4090 GPUs. Despite the computationally intensive diffusion process (1,000 steps), the dual-GPU parallelization enabled efficient inference: generating a single $512^{3}$ volume took approximately 2 hours and 16 minutes, while a $1024^{3}$ gigavoxel volume required about 11 hours and 25 minutes. It is worth noting that for this reconstruction task, we set the Classifier-Free Guidance (CFG) scale to $s=1$ (i.e., disabled). This configuration was chosen to prioritize fidelity to the physical distribution of the training data, although the model supports adjustable guidance for enhanced generalization (as demonstrated in \ref{CFG}).

The generated $1024^{3}$ sample corresponds to a physical volume of approximately $1.22 \times 10^{-8} m^{3}$, containing tens of thousands of pore-throat structures. We validated the model's performance across macroscopic continuity, microscopic morphology, and physical transport properties.

\subsubsection{Macroscopic Continuity and Morphological Consistency}

\begin{table}
\centering
\caption{Comparison of physical properties across different reconstruction scales. Note that permeability is analyzed separately in Figure \ref{fig:size_effect} via sub-volume partitioning.}
\label{tab:large_scale_metrics}
\resizebox{\textwidth}{!}{
\begin{tabular}{lccc}
\hline
\textbf{Sample Name} & \textbf{Dimensions (Voxels)} & \textbf{Porosity ($\phi$)} & \textbf{Connectivity (Largest Cluster \%)} \\ \hline
Ground Truth & $1000^{3}$ & 0.2672 & 99.43\% \\
Gen\_512 & $512^{3}$ & 0.2701 & 99.15\% \\
Gen\_1024 & $1024^{3}$ & 0.2725 & 99.49\% \\ \hline
\end{tabular}
}
\end{table}

\begin{figure}[pos=htbp]
    \centering
    \includegraphics[width=0.99\textwidth]{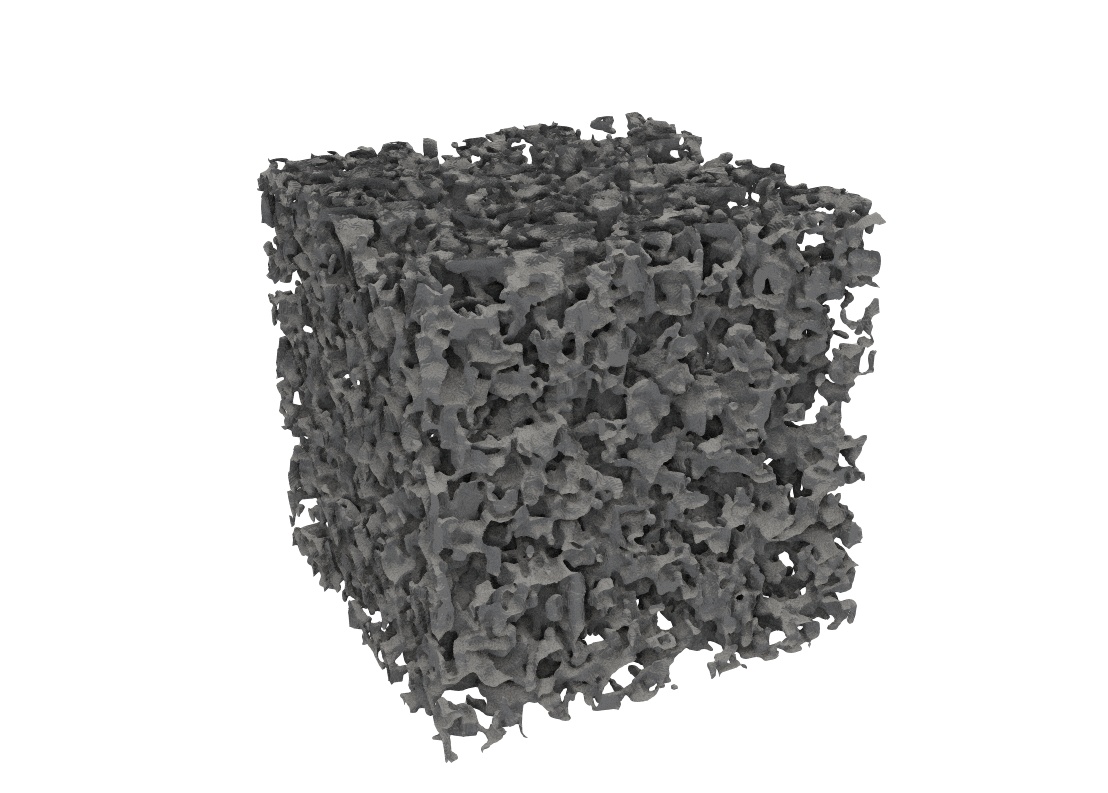} 
    \caption{Three-dimensional visualization of the pore network structure within the generated large-scale sample ($512^3$ voxels, since $1024^3$ is difficult to visualize).The grey isosurface represents the pore space, illustrating global connectivity without visible stitching artifacts.}
    \label{fig:large_scale_3d}
\end{figure}

\begin{figure}[pos=htbp]
    \centering
    \includegraphics[width=0.5\textwidth]{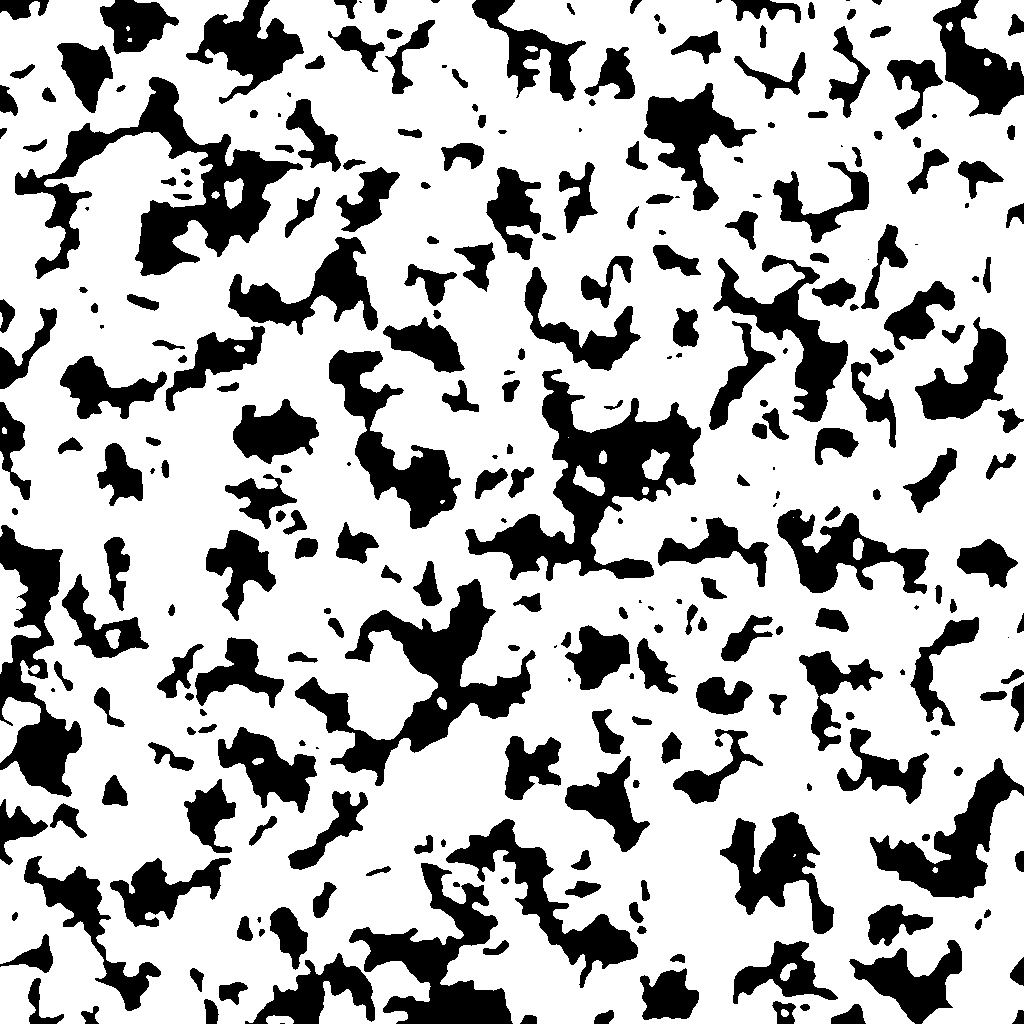}
    \caption{A representative orthogonal cross-section (Slice No. 0257) extracted from the $1024^3$ generated volume. Black regions denote pore space, and white regions denote the solid matrix. The image demonstrates seamless texture transition and long-range structural coherence.}
    \label{fig:large_scale_slice}
\end{figure}

We evaluated the effectiveness of the sliding-window stitching strategy by combining visual inspection with quantitative metrics listed in Table \ref{tab:large_scale_metrics}.

Visually, the isosurface rendering (Fig \ref{fig:large_scale_3d}) and the 2D cross-section (Fig \ref{fig:large_scale_slice}) demonstrate seamless macroscopic continuity. Pore channels extend naturally across stitching boundaries without any observable block artifacts. This qualitative observation is corroborated by the quantitative data in Table \ref{tab:large_scale_metrics}: the Largest Connected Cluster Fraction (Connectivity) for both $512^{3}$ and $1024^{3}$ samples remains consistently above 99\%, matching the Ground Truth. This confirms that the model preserves global topological integrity and avoids the formation of isolated "dead pores" often associated with patch-based generation.

\begin{figure}[pos=htbp]
    \centering
    \begin{subfigure}{0.48\textwidth}
        \includegraphics[width=\textwidth]{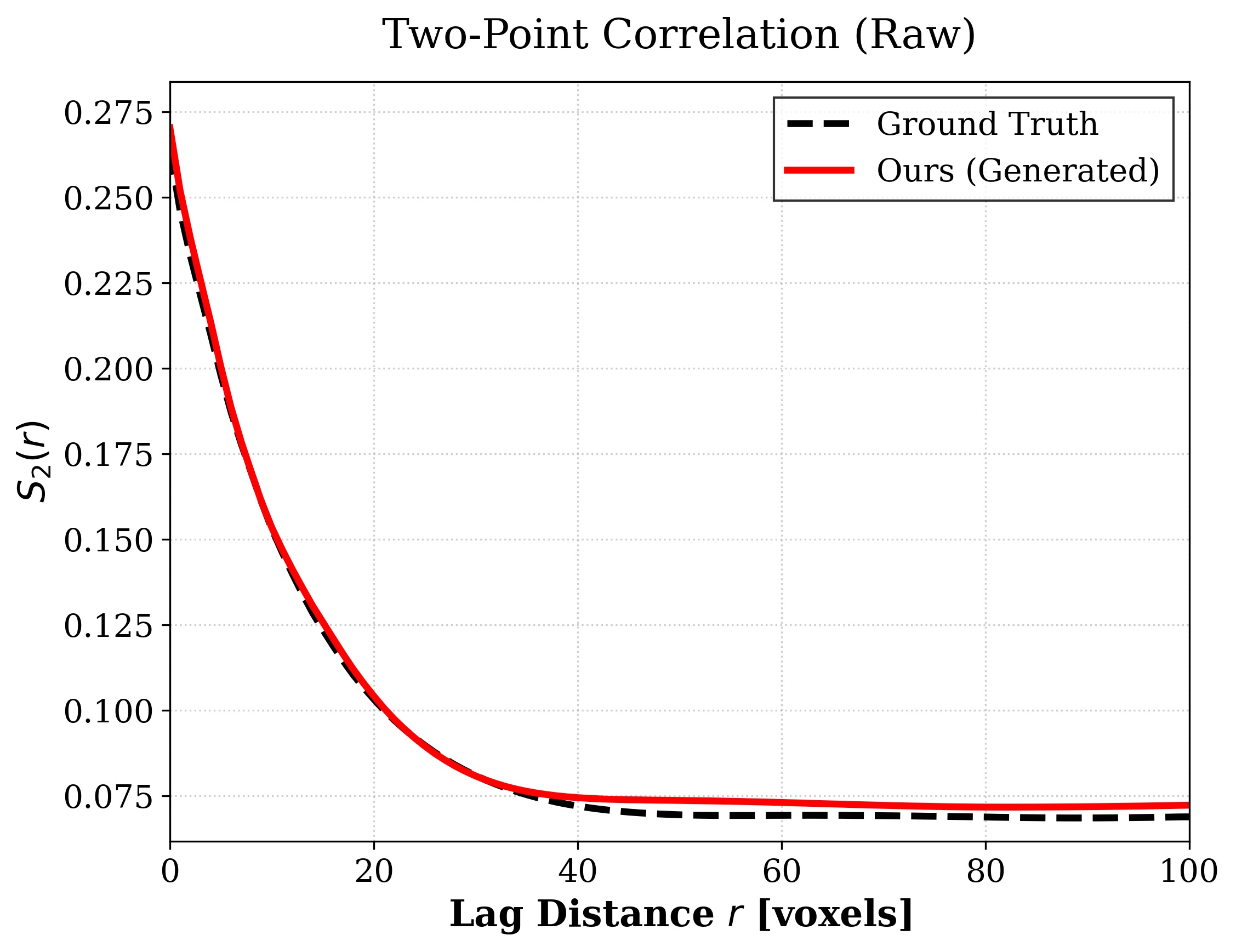}
        \caption{Two-point Correlation Function $S_2(r)$.}
        \label{fig:stats_s2}
    \end{subfigure}
    \hfill
    \begin{subfigure}{0.48\textwidth}
        \includegraphics[width=\textwidth]{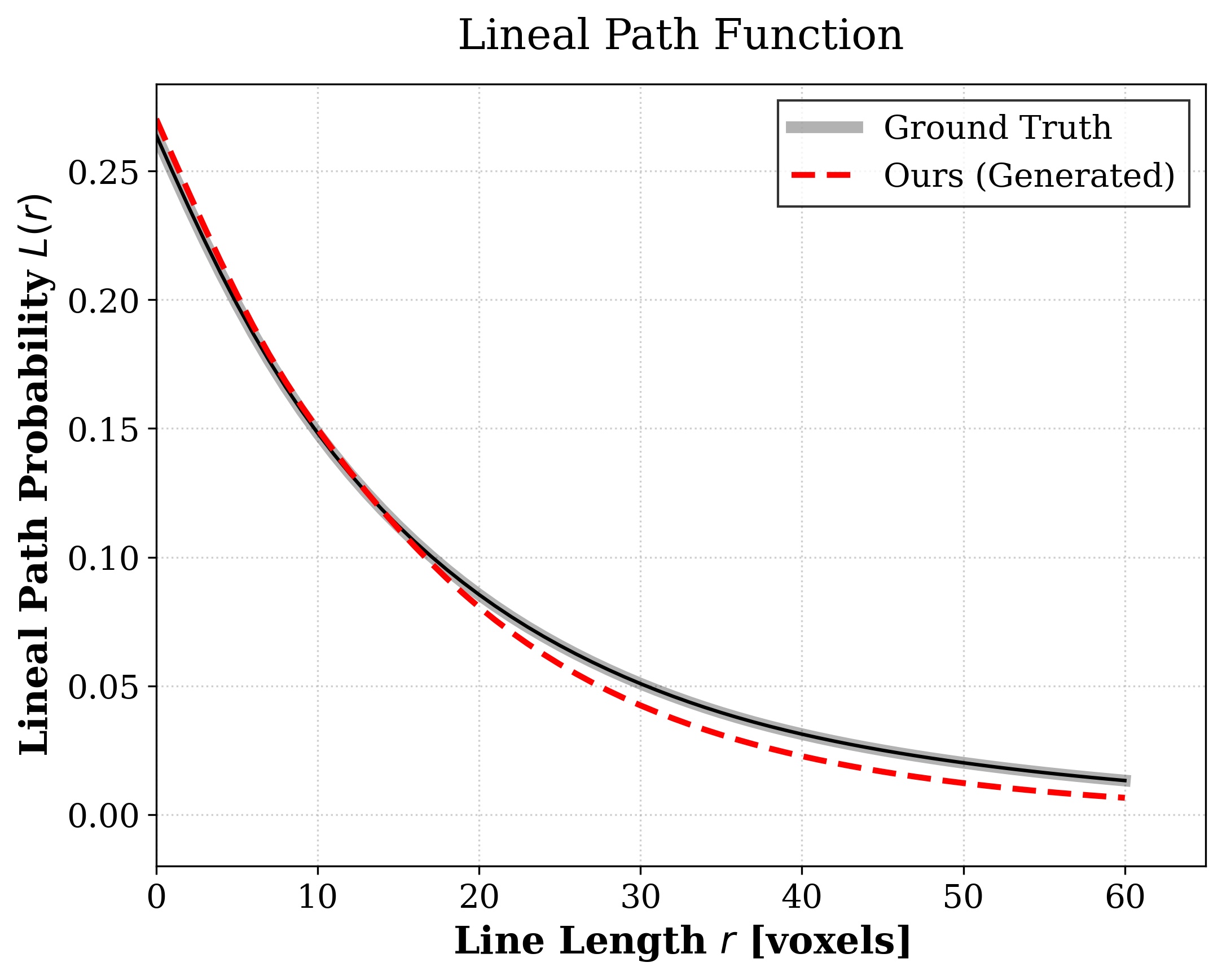}
        \caption{Lineal Path Probability $L(r)$.}
        \label{fig:stats_L}
    \end{subfigure}
    \caption{Statistical comparison between the Ground Truth (black dashed line) and the generated large-scale sample (red solid line). (a) The overlapping $S_2^*(r)$ curves indicate high statistical isomorphism in skeletal structure. (b) The parallel descent of the $L(r)$ curves suggests a consistent pore size distribution, despite a slight porosity offset.}
    \label{fig:large_scale_stats}
\end{figure}

In terms of microscopic morphology, Figure \ref{fig:large_scale_stats} compares the Two-point Correlation Function $S_{2}(r)$ and the Lineal Path Function $L(r)$. The curves of the generated samples overlap almost perfectly with those of the real rock, indicating that the model successfully reproduces the skeletal structure, characteristic correlation length, and pore size distribution during scale extrapolation. This statistical isomorphism verifies that PoreDiT can extrapolate physically valid macroscopic continua from local feature learning.

\subsubsection{Verification of Physical Properties Across Scales}

To assess fluid transport properties, we partitioned the generated $512^{3}$ and $1024^{3}$ volumes into 8 and 64 non-overlapping $256^{3}$ sub-volumes, respectively, and calculated their absolute permeability using LBM simulations (See more details in \ref{appendix_lbm}).
\begin{figure}[pos=htbp]
  \centering
  \includegraphics[width=0.6\linewidth]{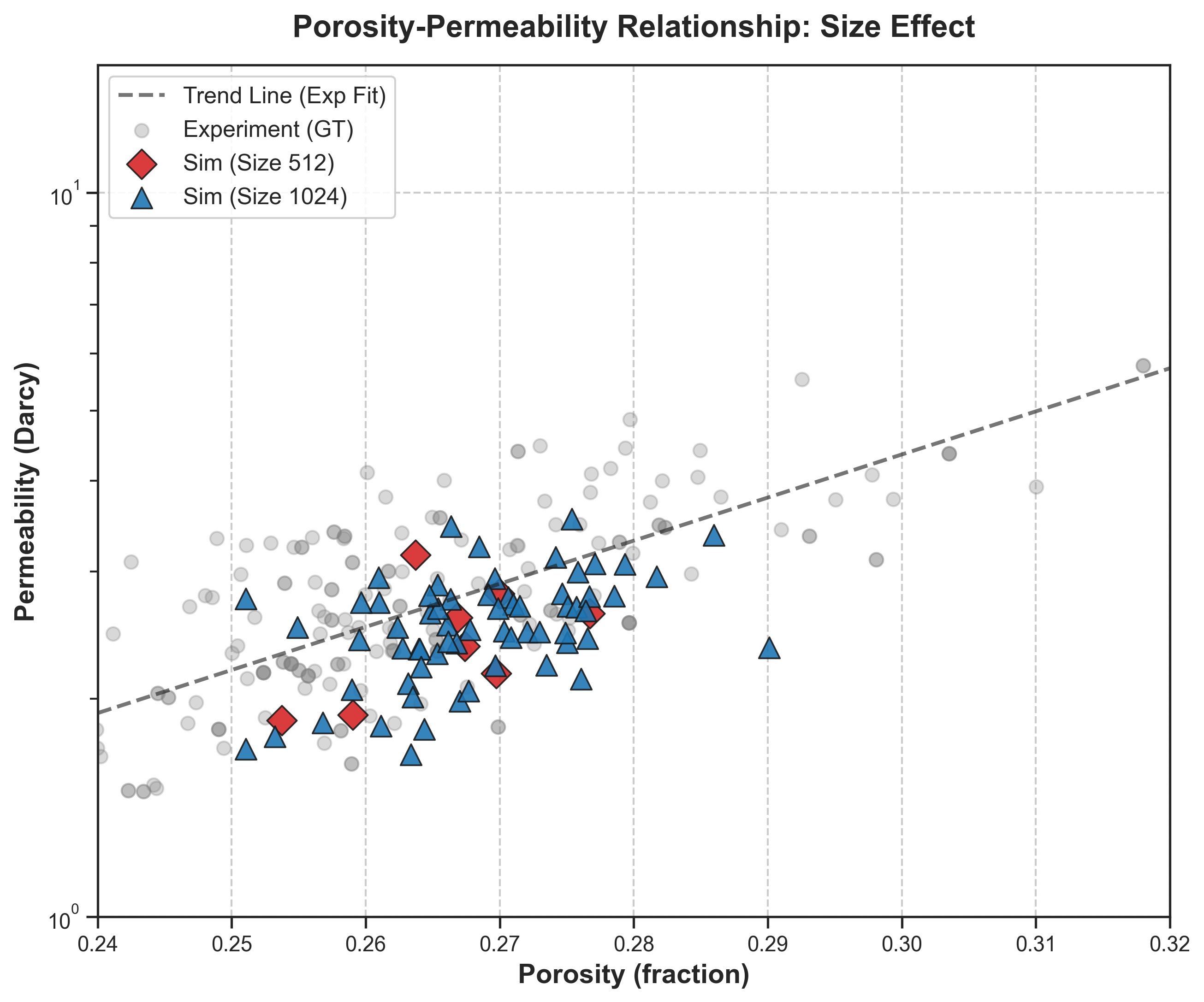}
  \caption{\textbf{Porosity-Permeability Relationship Analysis of Large-scale Generation.} The gray circles and dashed line represent the ground truth experimental data and its exponential fit trend. The red diamonds and blue triangles represent the sub-volumes sliced from the generated $512^3$ (8 sub-volumes) and $1024^3$ (64 sub-volumes) samples, respectively. The tight clustering around the GT trend line confirms physical consistency and stationarity across scales.}
  \label{fig:size_effect}
\end{figure}

Figure \ref{fig:size_effect} illustrates the joint porosity-permeability distribution. The data points from the generated sub-volumes cluster tightly around the experimental trend line of the real rock, showing no systematic bias. As shown in Table \ref{tab:large_scale_metrics}, the porosity of the generated samples (0.2701 and 0.2725) is highly consistent with the Ground Truth (0.2672). Furthermore, the permeability distribution falls entirely within the valid physical range. These results demonstrate that the model maintains local stationarity at the gigavoxel scale and accurately captures the complex coupling between pore geometry and fluid transport capacity.

\subsection{Generalization Verification: Experiments on Ketton Carbonate}
\label{ketton}
To validate the generalization capability of the proposed PoreDiT framework across distinct geological settings, we extended our experiments to reconstruct Ketton Limestone. 
Unlike the Bentheimer sandstone analyzed in the previous sections, Ketton is an oolitic carbonate rock. 
Its pore structure is governed by complex inter-granular contacts and cementation, exhibiting stronger heterogeneity and intricate topology. This distinct lithology presents a more significant challenge for generative modeling compared to clastic sandstones.

Addressing the structural complexity of carbonate rocks and adhering to computational constraints, we implemented two strategic refinements to the training methodology for this specific dataset:

\begin{enumerate}
    \item \textbf{Resolution Adjustment:} To optimize GPU memory usage while ensuring the capture of essential microscopic details, the input sub-volume resolution for training was adjusted to $192^3$ voxels.
    \item \textbf{Augmented Physical Conditioning:} Empirical observations indicated that using scalar porosity alone as a condition was insufficient to strictly constrain the complex pore-throat connectivity of carbonate samples. Consequently, we augmented the conditioning vector by explicitly incorporating statistical features derived from the \textit{Two-point Correlation Function} ($S_2$). This joint conditioning mechanism---integrating porosity with morphological statistics---provides more robust guidance, enabling the diffusion model to accurately learn the specific spatial texture and clustering patterns of the limestone.
\end{enumerate}

\begin{figure}[pos=htbp]
    \centering
    \begin{minipage}{0.49\linewidth}
        \centering
        \includegraphics[width=\linewidth]{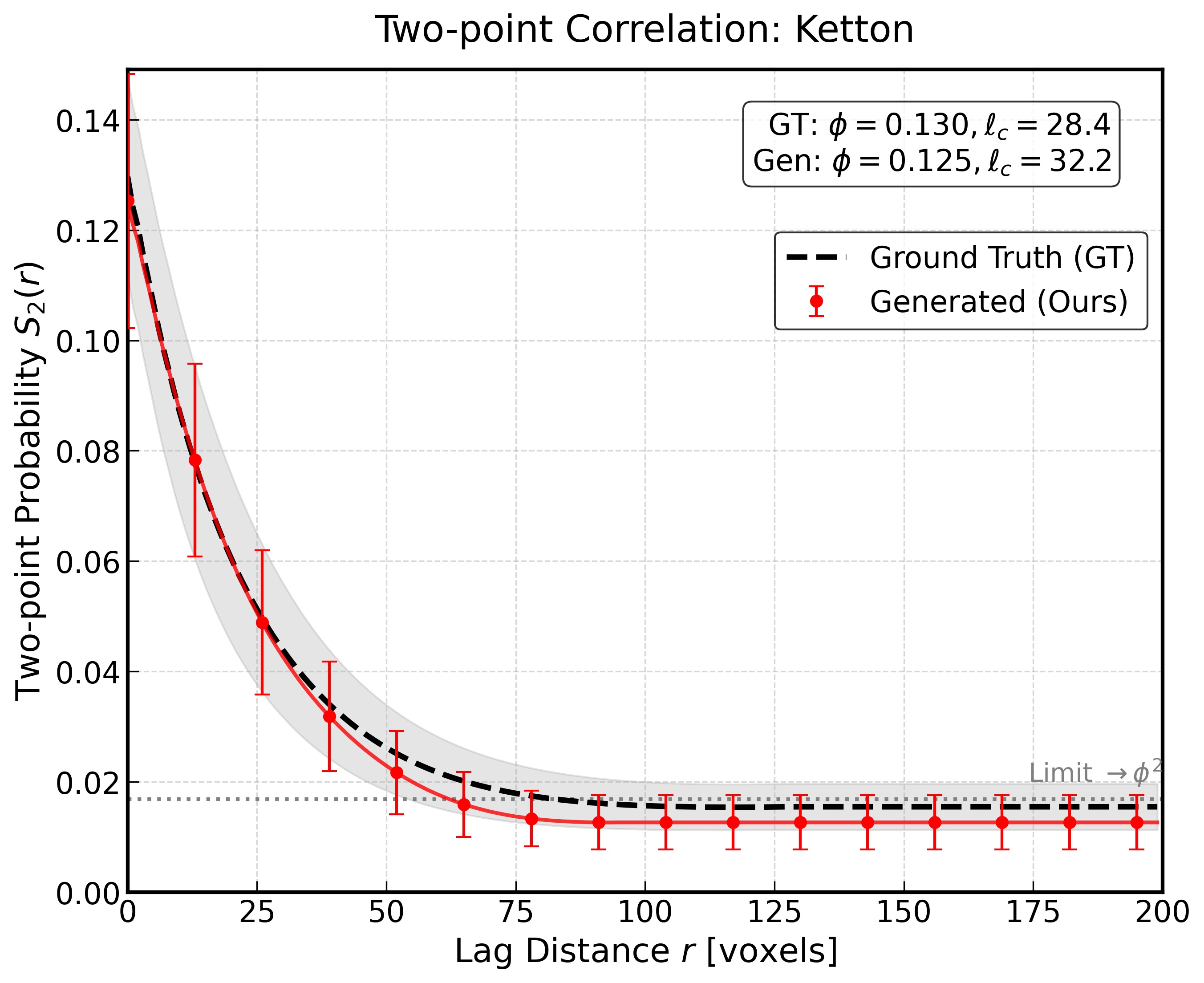}
        \centerline{(a)}
    \end{minipage}
    \hfill
    \begin{minipage}{0.49\linewidth}
        \centering
        \includegraphics[width=\linewidth]{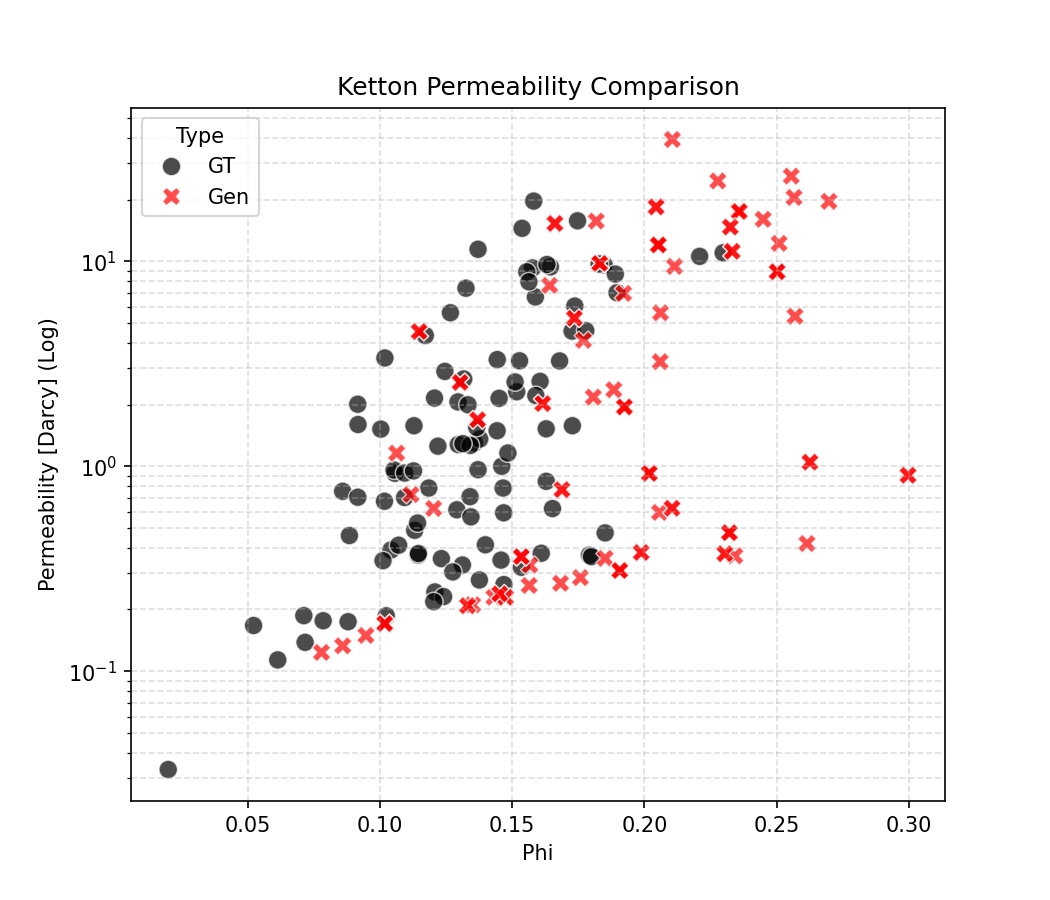}
        \centerline{(b)}
    \end{minipage}
    \caption{Generalization verification on Ketton Limestone. (a) Comparison of the Two-point Correlation Function $S_{2}(r)$. The generated samples (red) exhibit high morphological consistency with the ground truth (black), accurately reproducing the characteristic correlation length ($l_c$) and specific surface area. (b) Joint porosity-permeability distribution. The generated samples (red crosses) tightly cluster around the trend of the ground truth data (black circles), demonstrating that the model preserves the physical transport properties of the carbonate rock.}
    \label{fig:ketton_combined}
\end{figure}

Figure \ref{fig:ketton_combined} presents the quantitative validation results,using the same evaluation metrics established in Section~\ref{sec:morphological_analysis} and \ref{sec:permeability} The left panel (a) compares the Two-point Correlation Function ($S_2(r)$). Despite the structural complexity of the Ketton sample, the curve of the generated output (red solid line) aligns closely with the Ground Truth (black dashed line). The consistency in the slope at the origin (indicative of specific surface area) and the characteristic length ($l_c \approx 32.2$ voxels) confirms that the model successfully captures the granular packing characteristics inherent to oolitic limestone.

Furthermore, the right panel (b) illustrates the joint distribution of porosity and absolute permeability ($\phi-K$). The generated samples (red crosses) overlap extensively with the real data points (black circles). The results indicate that the synthetic samples not only match the target porosity distribution but also exhibit permeability values that adhere to the physical laws governing the real rock. These findings demonstrate that, with minor methodological adaptability regarding input conditioning, PoreDiT can effectively generalize to generate physically realistic digital cores for diverse and complex lithologies.

\subsection{Comprehensive Comparison with State-of-the-Art Methods}

Due to discrepancies in training datasets (e.g., Berea, Ketton, Carbonate) and evaluation metrics across the literature, a direct quantitative ranking presents challenges. Therefore, we provide a comprehensive qualitative comparison of representative methods in Table~\ref{tab:comparison}.

Traditional stochastic methods \citep{yeong1998reconstructing}, while foundational, often struggle to reproduce long-range connectivity. Early generative models represented by DCGAN \citep{mosser2017reconstruction} and WGAN-GP \citep{zha2020reconstruction} significantly improved inference speed but were often constrained to smaller generation sizes (e.g., $64^3$) due to GPU memory bottlenecks and training instability. Although PG-GAN \citep{you20213d} achieved large-scale reconstruction through progressive growing, its training cost is prohibitively high ($>200$ hours). Recent approaches such as RockGPT \citep{zheng2022rockgpt} and diffusion-based models \citep{naiff2025controlled} have made breakthroughs in physical fidelity but typically demand expensive computational clusters (e.g., 4--6 A100/V100 GPUs) and extended training periods.

In contrast, PoreDiT demonstrates superior comprehensive performance across three key dimensions: (1) \textbf{Topological and Physical Fidelity}: It achieves high accuracy in $S_2$ mismatch, permeability error, and connectivity ($>99\%$); (2) \textbf{VRAM Efficiency and Speed}: Training can be completed within 15 hours using consumer-grade GPUs (RTX 4090), significantly lowering the computational barrier; and (3) \textbf{High-Quality Large-Scale Generation}: It successfully realizes true 3D reconstruction at $1024^3$ resolution, overcoming the size limitations of existing diffusion models while maintaining structural precision.

\begin{table*}[pos=h]
    \centering
    \footnotesize 
    \setlength{\tabcolsep}{3pt} 
    \renewcommand{\arraystretch}{1.1} 

    \caption{Comparative analysis of PoreDiT against established reconstruction paradigms. The comparison is divided into two parts for clarity. \textbf{Part I} focuses on methodology and physical fidelity metrics ($S_2$, $K$, Connectivity), while \textbf{Part II} details scalability and computational efficiency metrics. Values for PoreDiT are experimentally derived from the Bentheimer sandstone dataset ($1024^3$).}
    \label{tab:comparison}

    \begin{tabular}{lllll} 
        \toprule
        \multicolumn{5}{c}{\textbf{Part I: Methodology and Physical Fidelity}} \\
        \midrule
        \textbf{Method} & \textbf{Paradigm} & \textbf{S$_2$ Mismatch} & \textbf{Permeability ($K$)} & \textbf{Connectivity} \\ 
        \midrule
        \textbf{Stochastic (SA)} & Optimization (SA) & Very Low & --- & Poor ($S_2$ only) / Good (Hybrid) \\ 
        \textbf{DCGAN} & Adversarial Training & Good Agreement & Good Match & Good (Euler Characteristic) \\ 
        \textbf{WGAN-GP} & Deep Learning & Good Agreement & --- & Qualitatively Good (No Metric) \\ 
        \textbf{PG-GAN} & Deep Learning & --- & --- & Good (Pore Network Matched) \\ 
        \textbf{RockGPT} & DL (Autoregressive) & Good Agreement & Good (Dist. Match) & Good (Euler Characteristic) \\ 
        \textbf{DDPM} & DL (Diffusion) & $<5.2\%$ & --- & Good (Lineal Path Verified) \\ 
        \textbf{Control-LDM} & DL (Diffusion) & Good Match & Satisfactory & Implicitly Good \\ 
        \textbf{PoreDiT (Ours)} & \textbf{DiT} & \textbf{2--4\%} & \textbf{$<$1\%} (at $512^3$) & \textbf{High ($>$99\%)} \\ 
        \bottomrule
    \end{tabular}

    \vspace{0.4cm} 

    \begin{tabular}{llll}
        \toprule
        \multicolumn{4}{c}{\textbf{Part II: Scalability and Computational Efficiency}} \\
        \midrule
        \textbf{Method} & \textbf{Scalability (Max Size)} & \textbf{Training Hardware} & \textbf{Training Time} \\ 
        \midrule
        \textbf{Stochastic (SA)} & Low (2D Only, $400^2$) & --- & --- \\ 
        \textbf{DCGAN} & High (Arbitrary Size) & NVIDIA K40 GPUs & $\sim 24$ Hours \\ 
        \textbf{WGAN-GP} & Low (Fixed $64\times64$) & NVIDIA GTX 1080Ti & $\sim 5$ Hours \\ 
        \textbf{PG-GAN} & High ($1024\times1024\times Z$) & NVIDIA Tesla V100 & 243 Hours \\ 
        \textbf{RockGPT} & Low ($64^3$ voxels) & 4 $\times$ NVIDIA Tesla V100 & $\sim 27.5$ Hours \\ 
        \textbf{DDPM} & Medium ($256 \times 256$, 2D) & NVIDIA RTX A6000 & --- \\ 
        \textbf{Control-LDM} & High ($256^3$ voxels) & 6 $\times$ NVIDIA A100 & 36--72 Hours \\ 
        \textbf{PoreDiT (Ours)} & \textbf{High ($1024^3$)} & \textbf{2 $\times$ NVIDIA RTX 4090} & \textbf{$\sim$ 15 Hours} \\ 
        \bottomrule
    \end{tabular}
\end{table*}

\section{Discussion}
\subsection{Mechanism Interpretation and Computational Efficiency}
PoreDiT achieves gigavoxel-scale ($1024^3$) extrapolation of digital rock cores on consumer-grade GPUs while maintaining high fidelity in microscopic pore structures. In contrast to traditional GAN-based approaches that rely on pseudo-3D processing, our method operates directly in three-dimensional space, effectively eliminating the mode collapse issue inherent to GANs. Furthermore, departing from the complex U-Net architectures prevalent in standard diffusion models, we utilize an isotropic 3D Swin Transformer backbone, which significantly reduces memory consumption during both training and inference.

The underlying reason for this exceptional computational efficiency and memory optimization lies in our effective utilization of the binary nature of porous media. Recognizing that digital rocks are characterized by a distinct pore-matrix duality (i.e., pixels are either black or white), we model the training process as an evolution within the probability field of the pixel space. Notably, PoreDiT employs an end-to-end design that directly predicts the clean original image $x_0$ rather than the noise, a design philosophy that coincides with the concurrent work JiT \citep{li2025back}. According to the Manifold Hypothesis, although digital rock data resides in a high-dimensional pixel space, its valid physical structures occupy a low-dimensional manifold. By leveraging binary features and adopting the $x_0$-prediction strategy, our model implicitly identifies and maps to this low-dimensional manifold within the high-dimensional space. This theoretical alignment explains why PoreDiT can efficiently capture complex pore topologies with a relatively streamlined architecture, thereby achieving state-of-the-art (SOTA) performance in generating ultra-large-scale ($1024^3$) volumes at speeds significantly surpassing existing methods.

\subsection{Paradigm Comparison with State-of-the-Art Methods}
Although a detailed quantitative comparison has been presented in the experimental section (see Table \ref{tab:comparison}), it is essential to revisit the methodological superiority of PoreDiT. Compared to traditional GAN-based approaches (e.g., WGAN-GP, PG-GAN), PoreDiT not only fundamentally eliminates the risk of mode collapse associated with unstable adversarial training but also achieves superior preservation of long-range connectivity by learning the evolution of probability density rather than direct texture mapping.

Specifically, in comparison with the current state-of-the-art (SOTA) work by Naiff et al.\cite{naiff2025controlled}, which employs an Elucidated Diffusion Model (EDM)-based Latent Diffusion Model to achieve high-fidelity reconstruction, a critical limitation remains. Naiff et al.\cite{naiff2025controlled} explicitly state that their generation volume is constrained to approximately $256^3$ voxels due to the prohibitive Video RAM (VRAM) consumption inherent to their 3D convolutional architecture. This architectural bottleneck prevents their method from scaling up to meet industrial demands for gigavoxel-level digital cores. In contrast, PoreDiT circumvents this ``memory wall'' by discarding the memory-intensive U-Net architecture in favor of an isotropic 3D Swin Transformer combined with a sliding-window extrapolation strategy. This design enables the seamless reconstruction of ultra-large-scale $1024^3$ volumes on a single consumer-grade GPU, demonstrating the superior scalability and engineering applicability of our architecture for massive 3D volumetric data.

\subsection{Limitations and Defense of Physical Validity}
\label{limitations}
Despite PoreDiT's exceptional performance in macroscopic statistics and transport properties, close visual inspection reveals that the generated pore edges appear slightly rounder compared to the original CT scans. We attribute this phenomenon to the ``implicit smoothing'' characteristic inherent in the diffusion denoising process. However, from the perspective of Computational Fluid Dynamics (CFD) simulation, this smoothing is not a defect but rather a beneficial feature. Excessively sharp or jagged edges in raw CT images often result from imaging noise or partial volume effects, which can complicate mesh generation. Smoother boundaries, in contrast, facilitate mesh convergence, thereby enhancing the stability of physical simulations.

Furthermore, we observe a loss of certain extremely minute pores in the generated samples, a consequence of high-frequency information compression during Patch Embedding. Physically, however, these micropores are typically isolated ``dead pores'' or scanning noise, with negligible contribution to fluid flow. The model successfully captures the connected backbone network that dominates permeability, reflecting the generative model's ability to focus on dominant modes. This also explains the slight deviation of approximately 0.8 Darcy in the permeability distribution (a conservative estimate): while the smoothing of pore throats slightly increases flow resistance, the predicted results remain entirely within the statistical distribution of the ground truth, confirming that the generated digital cores are physically robust and reliable.

\subsection{Practical Implications: Democratization of Digital Rock Physics}

The most significant practical contribution of PoreDiT lies in its drastic reduction of hardware barriers. Historically, generating gigavoxel-scale digital cores has relied heavily on expensive High-Performance Computing (HPC) clusters or multi-GPU parallelism. This study demonstrates that high-fidelity reconstruction at the $1024^3$ scale can be accomplished within hours using a single consumer-grade GPU (e.g., RTX 4090). This breakthrough significantly promotes the ``democratization'' of Digital Rock Physics (DRP): field engineers can now rapidly generate massive, diverse 3D pore structure datasets on local workstations without accessing supercomputers. These datasets can serve as direct inputs for subsequent pore-scale flow simulations (e.g., LBM or Pore Network Modeling (PNM)), thereby accelerating workflows for reservoir characterization and carbon sequestration feasibility analysis.

\subsection{Future Work}

Based on the limitations and potential of the current work, we propose three directions for future research:

\begin{enumerate}
    \item \textbf{Developing a ``Rock Zero'' Foundation Model}: Current deep learning approaches typically train dedicated models for specific rock samples (e.g., Bentheimer). However, in practical engineering, CT scan data for certain unique cores is often scarce. Inspired by Large Language Models (LLMs), we plan to collect massive datasets spanning diverse sources and lithologies to train an open-source, general-purpose foundation model. Engineers would then only need to fine-tune this pre-trained model with a small amount of target data to achieve high-quality reconstruction. This ``pre-training + fine-tuning'' paradigm holds promise for resolving modeling challenges in data-scarce scenarios.
    
    \item \textbf{Extension to Multiphase Probability Fields}: Currently, PoreDiT focuses on binary (pore/matrix) reconstruction. Future work will explore reconstruction involving multiple components, such as water, oil, and mineral matrices. We posit that as long as the phase states are limited and discrete, predicting within the pixel probability field (i.e., extending binary Sigmoid to multi-class Softmax) remains the theoretically optimal solution. This approach is expected to capture the topological relationships of phase interfaces more precisely than direct grayscale regression.
    
    \item \textbf{Generalization to Complex Lithologies}: Although we have validated the model's generalization on Ketton carbonate, further optimization is required for complex reservoirs with extreme heterogeneity, such as shales or rocks with micro-fracture networks. Future work will attempt to introduce stronger anisotropic prior constraints to adapt to a wider range of geological scenarios.
\end{enumerate}

\section{Conclusions}

In this work, we proposed  \textbf{PoreDiT}, a scalable diffusion-based generative framework designed to address a fundamental scalability bottleneck in porous microstructure reconstruction. By leveraging an isotropic 3D Swin Transformer architecture and directly modeling the binary probability field of pore space, the proposed method significantly reduces memory consumption while preserving long-range topological dependencies that are critical for physical transport processes.

A key contribution of this study is the demonstration that ultra-large porous microstructures, reaching the gigavoxel scale (up to $1024^3$ voxels), can be reconstructed efficiently on consumer-grade GPUs without relying on memory-intensive 3D convolutional backbones or high-performance computing clusters. This capability directly overcomes the long-standing trade-off between resolution, field-of-view, and computational cost that has constrained existing reconstruction approaches.

Extensive validation across multiple scales confirms that the generated microstructures exhibit strong physical fidelity. The reconstructed samples reproduce essential morphological, topological, and transport properties, including porosity, connectivity, correlation functions, and pore-scale permeability, with statistical consistency comparable to real materials and current state-of-the-art methods \citep{naiff2025controlled}. These results demonstrate that the proposed framework preserves not only visual realism but also physically meaningful structure–property relationships.

By substantially lowering the computational and hardware barriers associated with large-scale microstructure generation, this work facilitates the ``democratization'' of high-fidelity pore-scale modeling. It enables researchers to routinely generate massive, REV-compliant digital cores on local workstations, serving as direct inputs for advanced hydrodynamic simulations (e.g., LBM or PNM). This capability is particularly valuable for investigating scale-dependent flow mechanisms, such as viscous fingering, trapping efficiency in carbon sequestration, and enhanced oil recovery processes, establishing a new paradigm for data-driven computational fluid dynamics.

\section*{Data Availability Statement}
The Bentheimer sandstone dataset used in this study is available from the Digital Rocks Portal (Project DRP-317) \cite{neumann2020}. The Ketton Limestone dataset was obtained from the Imperial College London Pore-Scale Modelling and Imaging collection \cite{ImperialData, dong2009pore}. The source code implementing the PoreDiT framework, along with the pre-trained models and evaluation scripts, will be made openly available on GitHub at https://github.com/ruki-dot/PoreDiT upon publication.

\section*{Author Contributions}
\textbf{Yizhuo Huang} conducted the code development, performed the data analysis, and drafted the manuscript. \\
\textbf{Baoquan Sun} provided conceptual guidance for the study and contributed to the review and revision of the manuscript. \\
\textbf{Haibo Huang}, the corresponding author, supervised the overall research project and contributed to manuscript revision. \\
All authors reviewed and approved the final version of the manuscript.

\section*{Acknowledgments}
This research did not receive any specific grant from funding agencies in the public, commercial, or not-for-profit sectors.

\section*{Disclosure of interest}
The authors report there are no competing interests to declare.

\section*{Declaration of Generative AI and AI-assisted technologies in the writing process}
During the preparation of this work the author(s) used Gemini 3 Pro in order to improve the readability and language quality of the manuscript. After using this tool, the author(s) reviewed and edited the content and take(s) full responsibility for the content of the publication.

\appendix
\section{Supplementary Implementation Details}
\label{appendix:implementation_details}

This appendix provides detailed mathematical formulations and hyperparameter settings regarding the diffusion process discussed in Section 3.1.

\subsection{Noise Schedule and Frequency Embeddings}
\label{appendix:noise_schedule}

\textbf{Cosine Noise Schedule}\\
As introduced in Section 3.1.1, we employ a cosine schedule to optimize the signal-to-noise ratio decay. 
To prevent numerical singularities as $t \to 0$, the cumulative noise variance $\bar{\alpha}_t$ is defined as:

\begin{equation}
    \bar{\alpha}_{t} = \frac{f(t)}{f(0)}, \quad \text{where} \quad f(t) = \cos^2 \left( \frac{t/T + s}{1+s} \cdot \frac{\pi}{2} \right)
\end{equation}

Here, $T=1000$ represents the total diffusion steps. 
The offset $s$ is set to 0.008, ensuring that $\beta_t$ remains numerically stable near $t=0$ to stabilize the log-likelihood calculation during early training.

\textbf{Sinusoidal Frequency Embedding}\\
In Section 3.1.3, to map continuous scalars (time step $t$ and porosity $\phi$) into the high-dimensional latent space, we utilize sinusoidal frequency embeddings. 
Given a scalar input $s$ (representing either $t$ or $\phi$) and an embedding dimension $d=768$, the embedding vector for the $i$-th frequency component is calculated as:

\begin{equation}
    \begin{split}
    &\text{Emb}(s, 2i) = \sin\left(\frac{s}{10000^{2i/d}}\right), \\
    &\quad \text{Emb}(s, 2i+1) = \cos\left(\frac{s}{10000^{2i/d}}\right)
    \end{split}
\end{equation}

where $i \in [0, d/2 - 1]$. 
This embedding vector is subsequently processed by a Multi-Layer Perceptron (MLP) consisting of two fully connected layers with SiLU activation, projecting it to the condition vector $c$ that matches the Transformer's hidden dimension.

\subsection{Network Architecture Details}
\label{appendix:arch_details}

PoreDiT utilizes a Transformer-based backbone. 
To handle high-dimensional voxel data, the Patch Embedding layer is implemented using a \textbf{3D Convolution} (\texttt{Conv3d}) with both kernel size and stride set to 16. 
This configuration ensures non-overlapping linear projection. 
Detailed hyperparameters are listed in Table \ref{tab:hyperparameters}.

\begin{table*}
\centering
\caption{Hyperparameter Settings of PoreDiT}
\label{tab:hyperparameters}
\begin{tabular}{lc}
\toprule
\textbf{Hyperparameter} & \textbf{Value} \\
\midrule
Input Resolution & $256 \times 256 \times 256$ \\
Patch Size & $16 \times 16 \times 16$ \\
Embedding Dimension ($C$) & 768 \\
Depth ($L$) & 16 \\
Number of Heads & 12 \\
Window Size & $4 \times 4 \times 4$ \\
MLP Ratio & 4.0 \\
\bottomrule
\end{tabular}
\end{table*}

\subsection{Training Setup for Classifier-Free Guidance}
\label{appendix:cfg_training}

As mentioned in Section 3.1.3, to enable Classifier-Free Guidance (CFG) during inference, the model must learn the unconditional generation distribution during training. 
We achieve this via a \textbf{Conditional Dropout} mechanism.

During training, for each data batch, the porosity condition $c_{\phi}$ is randomly dropped with a probability of $p_{\text{drop}} = 0.1$. 
When dropped, the condition is replaced by a learnable null embedding vector $\varnothing \in \mathbb{R}^C$. 
Formally, the input condition $c_{\text{input}}$ is defined as:

\begin{equation}
    c_{\text{input}} = 
    \begin{cases} 
    \text{MLP}(\text{Emb}(\phi)), & \text{with probability } 1 - p_{\text{drop}} \\
    \varnothing, & \text{with probability } p_{\text{drop}}
    \end{cases}
\end{equation}

This training strategy enables the model to approximate both the conditional distribution $p_\theta(x|\phi)$ and the unconditional distribution $p_\theta(x)$, thereby allowing the trade-off between diversity and physical fidelity via the guidance scale during sampling.

\section{Attention Mechanism Details}
\label{appendix:transformer}

This appendix provides the mathematical formulation and complexity analysis of the Isotropic Swin-Transformer Encoder described in Section 3.2.1.

\subsection{3D Window Partitioning and W-MSA}

Let $z^{l-1} \in \mathbb{R}^{D \times H \times W \times C}$ denote the input feature map of the $l$-th block, where $D, H, W$ are spatial dimensions and $C$ is the channel dimension. The total number of tokens is $N = D \times H \times W$.
Standard global Multi-Head Self-Attention (MSA) requires computing relationships between all pairs of tokens, leading to a computational complexity of $\Omega(N^2)$, which is prohibitive for 3D volumetric data.

W-MSA (Window-based MSA) first partitions $z^{l-1}$ into non-overlapping local 3D windows of size $M \times M \times M$ (in this work, $M=4$). The total number of windows is $\frac{N}{M^3}$. For the flattened features $X_w \in \mathbb{R}^{M^3 \times C}$ within each local window, the Query ($Q$), Key ($K$), and Value ($V$) matrices are generated as:
\begin{equation}
    Q = X_w W^Q, \quad K = X_w W^K, \quad V = X_w W^V
\end{equation}
where $W^Q, W^K, W^V \in \mathbb{R}^{C \times d_k}$ are learnable projection matrices. To preserve spatial awareness, a relative position bias $B \in \mathbb{R}^{M^3 \times M^3}$ is added to the attention map. The attention within a window is computed as:
\begin{equation}
    \text{Attention}(Q, K, V) = \text{Softmax}\left(\frac{QK^T}{\sqrt{d_k}} + B\right)V
    \label{eq:window_attn}
\end{equation}

\subsection{Complexity Analysis: $O(N^2)$ vs $O(N)$}

Comparing the computational complexity (FLOPs) of standard global MSA and W-MSA for the entire volume:
\begin{itemize}
    \item \textbf{Global MSA}:
    \begin{equation}
        \Omega(\text{Global}) = 4NC^2 + 2N^2C
    \end{equation}
    The quadratic term $2N^2C$ makes global attention computationally intractable as $N$ scales cubically with volume resolution.
    
    \item \textbf{W-MSA}:
    \begin{equation}
        \Omega(\text{Window}) = 4NC^2 + 2N M^3 C
    \end{equation}
\end{itemize}
Since the window size $M$ is fixed (e.g., $M=4$), the term $M^3$ is a constant and $M^3 \ll N$. Consequently, the complexity of W-MSA becomes linear with respect to $N$, i.e., $O(N)$, enabling the efficient processing of high-resolution 3D porous media.

\subsection{Shifted Window Mechanism (SW-MSA)}

While efficient, W-MSA lacks connections across windows. SW-MSA addresses this in the $(l+1)$-th layer by shifting the window partitioning configuration by $(\lfloor \frac{M}{2} \rfloor, \lfloor \frac{M}{2} \rfloor, \lfloor \frac{M}{2} \rfloor)$ voxels along the $D, H, W$ axes.
To maintain computational efficiency (avoiding padding that would increase the number of windows), we employ a \textbf{cyclic shift} strategy. The feature map is cyclically shifted such that non-adjacent sub-windows are composed into a single batch window. A masking matrix $\mathcal{M}$ is introduced to mask out self-attention computations between these logically non-adjacent regions:
\begin{equation}
    \begin{split}
    &\text{MaskedAttention}(Q, K, V) = \\
    &\text{Softmax}\left(\frac{QK^T}{\sqrt{d_k}} + B + \mathcal{M}\right)V
    \end{split}
\end{equation}
This mechanism achieves cross-window global information exchange without increasing computational cost.

\section{Decoder Details and GELU Activation}
\label{appendix:decoder}

This appendix details the internal computation mechanisms of the asymmetric decoder (Final Layer) described in Section 3.2.2.
The inputs to the decoder are $Z \in \mathbb{R}^{N \times C}$ and the condition embedding vector $c$.

\subsection{Adaptive Layer Normalization (AdaLN)}
Unlike standard layer normalization, AdaLN dynamically regresses scale factor $\gamma_c$ and shift factor $\beta_c$ from the condition $c$:
\begin{equation}
    [\gamma_c, \beta_c] = \text{MLP}(c)
\end{equation}
The normalized feature $\hat{z}$ is computed as:
\begin{equation}
    \hat{z} = \gamma_c \cdot \text{LayerNorm}(Z) + \beta_c
\end{equation}
This step ensures that physical conditions (e.g., porosity) directly modulate the feature distribution.

\subsection{GELU Activation}
Before entering the linear layer, features are activated by the Gaussian Error Linear Unit (GELU). Compared to ReLU, GELU provides smooth non-linearity around zero, facilitating stable gradient propagation:
\begin{equation}
    \text{GELU}(x) = x \Phi(x) \approx 0.5x(1 + \tanh(\sqrt{2/\pi}(x + 0.044715x^3)))
\end{equation}

\subsection{Linear Projection and Unpatchify}
Finally, the activated features are projected and reshaped into voxel blocks:
\begin{equation}
    \hat{V} = \text{GELU}(\hat{z}) \mathbf{W}_{\text{out}}^T + \mathbf{b}_{\text{out}}
\end{equation}
\begin{equation}
    \text{Logits} = \text{Unpatchify}(\hat{V}) \in \mathbb{R}^{D \times H \times W}
\end{equation}

\section{Loss Function Details}
\label{appendix:loss}

This appendix provides detailed formulations for the loss components described in Section \ref{sec:loss}.

\subsection{Reconstruction Loss}
The reconstruction loss is a weighted sum of Binary Cross-Entropy (BCE) and Dice Loss:
\begin{equation}
    \mathcal{L}_{\text{rec}} = \mathcal{L}_{\text{BCE}} + \mathcal{L}_{\text{Dice}}
\end{equation}
For a predicted probability map $P$ and ground truth binary mask $G$:
\begin{equation}
    \mathcal{L}_{\text{BCE}} = - \frac{1}{N} \sum_i [G_i \log P_i + (1-G_i) \log (1-P_i)]
\end{equation}
\begin{equation}
    \mathcal{L}_{\text{Dice}} = 1 - \frac{2 \sum_i P_i G_i + \epsilon}{\sum_i P_i + \sum_i G_i + \epsilon}
\end{equation}
where $\epsilon$ is a smoothing term to prevent division by zero.

\subsection{Two-Point Correlation Function ($S_2$) Calculation}
The two-point correlation function $S_2(r)$ is efficiently computed on the probability field using spatial shifts. For a given lag vector $r$ along a specific axis (e.g., x-axis), the estimated $S_2$ value is:
\begin{equation}
    \hat{S}_2(r) = \frac{1}{|\Omega_r|} \sum_{x \in \Omega_r} P(x) \cdot P(x+r)
\end{equation}
where $\Omega_r$ represents the valid spatial domain for the shift. In our implementation, we calculate the average $S_2$ along the three Cartesian axes ($D, H, W$) for a set of lag distances $\mathcal{R} = \{8, 16, 32, 64, 96, 128\}$ voxels, ensuring the model captures spatial correlations at multiple scales.

\section{Inference Details and Variance Analysis}
\label{appendix:inference_details}

This appendix supplements Section \ref{sec:inference} with the exact sampling formulation used in our code and the mathematical derivation of the Global Coherent Noise strategy.

\subsection{Posterior Mean Calculation ($x_0$-Parameterization)}
Since our model predicts $\hat{x}_0$ directly, the reverse diffusion step relies on the posterior mean of the distribution $q(x_{t-1} | x_t, x_0)$. According to the DDPM schedule, this is computed as:
$$\tilde{\mu}_t = \left( \frac{\sqrt{\bar{\alpha}_{t-1}} \beta_t}{1 - \bar{\alpha}_t} \right) \cdot \hat{x}_0 + \left( \frac{\sqrt{\alpha_t} (1 - \bar{\alpha}_{t-1})}{1 - \bar{\alpha}_t} \right) \cdot x_t$$$$\tilde{\sigma}_t^2 = \frac{1 - \bar{\alpha}_{t-1}}{1 - \bar{\alpha}_t} \cdot \beta_t$$

where $\alpha_t, \bar{\alpha}_t, \beta_t$ are the standard noise schedule parameters. The next state $x_{t-1}$ is sampled as $x_{t-1} = \mu_{t-1} + \sigma_t z$, where $z$ is the noise term. This formulation ensures that the reconstruction is strictly guided by the predicted signal rather than the noise residual.

\subsection{Variance Collapse and Porosity Drift in Tiled Diffusion}
To mathematically explain why independent noise sampling leads to porosity drift, consider a voxel location covered by $N$ overlapping tiles. Let $x_1, x_2, \dots, x_N$ be the noise samples from these $N$ tiles at this location. During fusion, the voxel value is typically derived via weighted averaging:
\begin{equation}
    x_{\text{fused}} = \frac{\sum_{i=1}^N w_i x_i}{\sum_{i=1}^N w_i}
\end{equation}
For analytical clarity, we assume equal weights, simplifying to $x_{\text{fused}} = \frac{1}{N} \sum_{i=1}^N x_i$.

\textbf{Case 1: Independent Noise.}
If noise is sampled independently for each tile, i.e., $x_i \overset{i.i.d.}{\sim} \mathcal{N}(0, 1)$, the variance of the fused voxel becomes:
\begin{equation}
    \begin{split}
    &\text{Var}(x_{\text{fused}}) = \text{Var}\left(\frac{1}{N} \sum_{i=1}^N x_i\right) \\
    &= \frac{1}{N^2} \sum_{i=1}^N \text{Var}(x_i) = \frac{1}{N^2} \cdot N \cdot 1 = \frac{1}{N}
    \end{split}
\end{equation}
This implies that the input variance fluctuates drastically depending on the degree of overlap. In corner overlap regions ($N=8$), the variance collapses to $0.125$, while in face overlap regions ($N=2$), it drops to $0.5$. This non-uniformity and significant reduction in variance (far below the unit variance of 1) are misinterpreted by the model as variations in signal strength, leading to systematic porosity drift artifacts.

\textbf{Case 2: Global Coherent Noise.}
By initializing a global noise field, we enforce $x_1 = x_2 = \dots = x_N = x_{\text{global}}$. Consequently:
\begin{equation}
    \text{Var}(x_{\text{fused}}) = \text{Var}\left(\frac{1}{N} \sum_{i=1}^N x_{\text{global}}\right) = \text{Var}(x_{\text{global}}) = 1
\end{equation}
This strategy mathematically guarantees that variance remains constant (stationary) across the entire field, regardless of the overlap degree $N$.

\textbf{Impact on Porosity Drift:}
The diffusion model $f_\theta(x_t)$ is trained under the assumption that inputs follow a standard normal distribution $x_t \sim \mathcal{N}(0, 1)$. 
When variance collapses ($\text{Var} \ll 1$), the signal amplitude received by the network is significantly lower than expected. The model misinterprets this as a different signal-to-noise ratio, tending to output conservative predictions close to the mean (gray values) rather than high-contrast binary (pore/grain) structures. Upon final thresholding, this tendency towards ``grayness'' manifests as a systematic bias in the number of pore voxels, termed \textbf{Porosity Drift}. Global Coherent Noise preserves the correct input variance, ensuring the model generates binary structures with the correct statistical distribution within every sliding window.

\section{Calculation of Physical Descriptors}
\label{appendix_metrics}

To ensure the reproducibility of our statistical analysis, this appendix details the mathematical definitions and computational implementations of the descriptors used in Section \ref{sec:morphological_analysis}. All calculations were performed using the \texttt{PoreSpy} \citep{gostick2019porespy} and \texttt{scikit-image} libraries.

\subsection{Two-point Correlation Function ($S_2$)}
\label{appendix_s2}
The two-point correlation function $S_2(\mathbf{r})$ is defined as the probability that two points separated by a vector $\mathbf{r}$ both lie within the pore phase $\mathcal{P}$. For isotropic media such as Bentheimer sandstone, this simplifies to a function of the scalar distance $r = |\mathbf{r}|$:
\begin{equation}
    S_2(r) = \langle \mathcal{I}(\mathbf{x}) \mathcal{I}(\mathbf{x} + \mathbf{r}) \rangle
\end{equation}
where $\mathcal{I}(\mathbf{x})$ is the indicator function, equal to 1 if $\mathbf{x} \in \mathcal{P}$ and 0 otherwise, and $\langle \cdot \rangle$ denotes volume averaging. We computed $S_2(r)$ using the Fast Fourier Transform (FFT) method, which accelerates the convolution operation. The physical significance of $S_2$ is twofold: its limit at infinity relates to porosity ($\lim_{r \to \infty} S_2(r) = \phi^2$), and its slope at the origin relates to the specific surface area $S_v$ \citep{torquato2002random}:
\begin{equation}
    \frac{dS_2}{dr}\bigg|_{r=0} = -\frac{S_v}{4}
\end{equation}

\subsection{Minkowski Functionals and Topological Cleaning}
\label{appendix_minkowski}
Minkowski functionals provide a comprehensive characterization of the morphology and topology of binary structures. In three-dimensional Euclidean space, the four functionals correspond to Volume ($V_0$), Surface Area ($V_1$), Mean Curvature ($V_2$), and Euler Characteristic ($V_3$, often denoted as $\chi$). In our statistical analysis, we focused on:
\begin{equation}
    \begin{split}
    &M_0(K) = V(K), \quad M_1(K) = \int_{\partial K} dS, \\
    &\quad M_2(K) = \int_{\partial K} \frac{1}{2}(\kappa_1 + \kappa_2) dS
    \end{split}
\end{equation}
where $\kappa_1$ and $\kappa_2$ are the principal curvatures of the pore surface $\partial K$.

The Euler Characteristic ($\chi \propto M_2$) is a topological invariant that describes the connectivity of the pore network:
\begin{equation}
    \begin{split}
    &\chi = (\text{number of isolated objects}) - (\text{number of tunnels}) \\
    &+ (\text{number of cavities})
    \end{split}
\end{equation}
In X-ray CT imaging and voxel-based generation, single-pixel noise or scanning artifacts can manifest as tiny, non-physical isolated clusters, which disproportionately skew the Euler number. To mitigate this discretization error, we applied a strict \textbf{morphological cleaning} step before calculating topological metrics. Based on the resolution analysis by Cnudde et al.\cite{cnudde2013high}, we defined a physical validity threshold corresponding to a sphere with a diameter of 4 voxels. Any isolated pore cluster with a volume $V < \frac{\pi}{6}(4)^3 \approx 34$ voxels was identified as noise and removed. This preprocessing was consistently applied to both Ground Truth and generated samples to ensure a fair comparison of the macroscopic topology.

\section{Lattice Boltzmann Simulation Details}
\label{appendix_lbm}

To ensure the reproducibility of our permeability calculations, this appendix details the implementation of the Lattice Boltzmann Method (LBM) used in Section~\ref{sec:permeability}.

\subsection{Governing Equations and Discretization}
In this study, we employed the Bhatnagar-Gross-Krook (BGK) collision operator \citep{BGK1954} with the standard D3Q19 (three dimensions, nineteen velocities) lattice arrangement \citep{qian1992lattice}. The evolution of the particle distribution function is governed by the discrete lattice Boltzmann equation:
\begin{equation}
    f_i(\mathbf{x} + \mathbf{e}_i \Delta t, t + \Delta t) - f_i(\mathbf{x}, t) = -\frac{1}{\tau} [f_i(\mathbf{x}, t) - f_i^{eq}(\mathbf{x}, t)]
\end{equation}
where $f_i$ denotes the particle distribution function, $\mathbf{e}_i$ represents the discrete velocity vectors, and $\tau$ is the dimensionless relaxation time related to the kinematic viscosity $\nu$.

\subsection{Boundary Conditions and Permeability Calculation}
For the solid-fluid interface (pore walls), the standard \textbf{half-way bounce-back} scheme \citep{ladd1994numerical} was applied to enforce the no-slip boundary condition. To drive the fluid flow, pressure boundary conditions \citep{zou1997pressure} were imposed at the inlet and outlet, establishing a constant pressure gradient $\nabla P$ along the flow direction.

The simulation was considered converged when the relative error of the velocity field fell below a tolerance of $10^{-5}$. Upon convergence, the absolute permeability $K$ was calculated using Darcy's Law:
\begin{equation}
    K = \frac{\mu \bar{u} L}{\Delta P}
\end{equation}
where $\bar{u}$ is the average superficial velocity (Darcy velocity) across the domain, $L$ is the domain length, and $\mu$ is the dynamic viscosity.

\bibliographystyle{elsarticle-num} 
\bibliography{bibliography}

@article{torquato1993chord,
  title={Chord-length distribution function for two-phase random media},
  author={Torquato, Salvatore and Lu, B},
  journal={Physical Review E},
  volume={47},
  number={4},
  pages={2950},
  year={1993},
  publisher={APS}
}

@article{hazlett1997statistical,
  title={Statistical characterization and stochastic modeling of pore networks in relation to fluid flow},
  author={Hazlett, RD},
  journal={Mathematical Geology},
  volume={29},
  number={6},
  pages={801--822},
  year={1997},
  publisher={Springer}
}

@article{zhu2019challenges,
  title={Challenges and prospects of digital core-reconstruction research},
  author={Zhu, Linqi and Zhang, Chong and Zhang, Chaomo and Zhou, Xueqing and Zhang, Zhansong and Nie, Xin and Liu, Weinan and Zhu, Boyuan},
  journal={Geofluids},
  volume={2019},
  number={1},
  pages={7814180},
  year={2019},
  publisher={Wiley Online Library}
}

@article{mosser2017reconstruction,
  title={Reconstruction of three-dimensional porous media using generative adversarial neural networks},
  author={Mosser, Lukas and Dubrule, Olivier and Blunt, Martin J},
  journal={Physical Review E},
  volume={96},
  number={4},
  pages={043309},
  year={2017},
  publisher={APS}
}

@article{zha2020reconstruction,
  title={Reconstruction of shale image based on Wasserstein Generative Adversarial Networks with gradient penalty},
  author={Zha, Wenshu and Li, Xingbao and Xing, Yan and He, Lei and Li, Daolun},
  journal={Advances in Geo-Energy Research},
  volume={4},
  number={1},
  pages={107--114},
  year={2020}
}

@article{you20213d,
  title={3D carbonate digital rock reconstruction using progressive growing GAN},
  author={You, Nan and Li, Yunyue Elita and Cheng, Arthur},
  journal={Journal of Geophysical Research: Solid Earth},
  volume={126},
  number={5},
  pages={e2021JB021687},
  year={2021},
  publisher={Wiley Online Library}
}

@article{zhu2024generation,
  title={Generation of pore-space images using improved pyramid Wasserstein generative adversarial networks},
  author={Zhu, Linqi and Bijeljic, Branko and Blunt, Martin J},
  journal={Advances in Water Resources},
  volume={190},
  pages={104748},
  year={2024},
  publisher={Elsevier}
}

@article{dosovitskiy2020image,
  title={An image is worth 16x16 words: Transformers for image recognition at scale},
  author={Dosovitskiy, Alexey},
  journal={arXiv preprint arXiv:2010.11929},
  year={2020}
}

@article{zheng2022rockgpt,
  title={RockGPT: reconstructing three-dimensional digital rocks from single two-dimensional slice with deep learning},
  author={Zheng, Qiang and Zhang, Dongxiao},
  journal={Computational Geosciences},
  volume={26},
  number={3},
  pages={677--696},
  year={2022},
  publisher={Springer}
}

@inproceedings{ma2023enhancing,
  title={Enhancing the resolution of micro-CT images of rock samples via unsupervised machine learning based on a diffusion model},
  author={Ma, Zhaoyang and Sun, Shuyu and Yan, Bicheng and Kwak, Hyung and Gao, Jun},
  booktitle={SPE Annual Technical Conference and Exhibition},
  pages={D021S028R005},
  year={2023},
  organization={SPE}
}

@article{vlassis2023denoising,
  title={Denoising diffusion algorithm for inverse design of microstructures with fine-tuned nonlinear material properties},
  author={Vlassis, Nikolaos N and Sun, WaiChing},
  journal={Computer Methods in Applied Mechanics and Engineering},
  volume={413},
  pages={116126},
  year={2023},
  publisher={Elsevier}
}

@article{park2024inverse,
  title={Inverse design of porous materials: a diffusion model approach},
  author={Park, Junkil and Gill, Aseem Partap Singh and Moosavi, Seyed Mohamad and Kim, Jihan},
  journal={Journal of Materials Chemistry A},
  volume={12},
  number={11},
  pages={6507--6514},
  year={2024},
  publisher={Royal Society of Chemistry}
}

@article{naiff2025controlled,
  title={Controlled Latent Diffusion Models for 3D Porous Media Reconstruction},
  author={Naiff, Danilo and Schaeffer, Bernardo P and Pires, Gustavo and Stojkovic, Dragan and Rapstine, Thomas and Ramos, Fabio},
  journal={arXiv preprint arXiv:2503.24083},
  year={2025}
}

@article{li2025back,
  title={Back to basics: Let denoising generative models denoise},
  author={Li, Tianhong and He, Kaiming},
  journal={arXiv preprint arXiv:2511.13720},
  year={2025}
}

@article{Qian1992Lattice,
  title={Lattice BGK models for Navier-Stokes equation},
  author={Qian, Y. H. and d'Humi{\`e}res, D. and Lallemand, P.},
  journal={Europhysics Letters},
  volume={17},
  number={6},
  pages={479--484},
  year={1992},
  publisher={IOP Publishing}
}

@article{BGK1954,
  title={A model for collision processes in gases. I. Small amplitude processes in charged and neutral one-component systems},
  author={Bhatnagar, Prabhu L and Gross, Eugene P and Krook, Max},
  journal={Physical review},
  volume={94},
  number={3},
  pages={511},
  year={1954},
  publisher={APS}
}

@book{Succi2001,
  title={The lattice Boltzmann equation: for fluid dynamics and beyond},
  author={Succi, Sauro},
  year={2001},
  publisher={Oxford university press}
}

@article{yeong1998reconstructing,
  title={Reconstructing random media},
  author={Yeong, Christofer LY and Torquato, Salvatore},
  journal={Physical review E},
  volume={57},
  number={1},
  pages={495},
  year={1998},
  publisher={APS}
}

@article{blunt2013pore,
  title={Pore-scale imaging and modelling},
  author={Blunt, Martin J and Bijeljic, Branko and Dong, Hu and Gharbi, Oussama and Iglauer, Stefan and Mostaghimi, Peyman and Paluszny, Adriana and Pentland, Christopher},
  journal={Advances in Water resources},
  volume={51},
  pages={197--216},
  year={2013},
  publisher={Elsevier}
}

@article{cnudde2013high,
  title={High-resolution X-ray computed tomography in geosciences: A review of the current technology and applications},
  author={Cnudde, Veerle and Boone, Matthieu Nicolaas},
  journal={Earth-Science Reviews},
  volume={123},
  pages={1--17},
  year={2013},
  publisher={Elsevier}
}

@article{bultreys2016imaging,
  title={Imaging and image-based fluid transport modeling at the pore scale in geological materials: A practical introduction to the current state-of-the-art},
  author={Bultreys, Tom and De Boever, Wesley and Cnudde, Veerle},
  journal={Earth-Science Reviews},
  volume={155},
  pages={93--128},
  year={2016},
  publisher={Elsevier}
}

@book{bear2013dynamics,
  title={Dynamics of fluids in porous media},
  author={Bear, Jacob},
  year={2013},
  publisher={Courier Corporation}
}

@article{ho2020denoising,
  title={Denoising diffusion probabilistic models},
  author={Ho, Jonathan and Jain, Ajay and Abbeel, Pieter},
  journal={Advances in neural information processing systems},
  volume={33},
  pages={6840--6851},
  year={2020}
}

@inproceedings{nichol2021improved,
  title={Improved denoising diffusion probabilistic models},
  author={Nichol, Alexander Quinn and Dhariwal, Prafulla},
  booktitle={International conference on machine learning},
  pages={8162--8171},
  year={2021},
  organization={PMLR}
}

@inproceedings{liu2021swin,
  title={Swin transformer: Hierarchical vision transformer using shifted windows},
  author={Liu, Ze and Lin, Yutong and Cao, Yue and Hu, Han and Wei, Yixuan and Zhang, Zheng and Lin, Stephen and Guo, Baining},
  booktitle={Proceedings of the IEEE/CVF international conference on computer vision},
  pages={10012--10022},
  year={2021}
}

@inproceedings{peebles2023scalable,
  title={Scalable diffusion models with transformers},
  author={Peebles, William and Xie, Saining},
  booktitle={Proceedings of the IEEE/CVF international conference on computer vision},
  pages={4195--4205},
  year={2023}
}

@article{ho2022classifier,
  title={Classifier-free diffusion guidance},
  author={Ho, Jonathan and Salimans, Tim},
  journal={arXiv preprint arXiv:2207.12598},
  year={2022}
}

@inproceedings{liu2022video,
  title={Video swin transformer},
  author={Liu, Ze and Ning, Jia and Cao, Yue and Wei, Yixuan and Zhang, Zheng and Lin, Stephen and Hu, Han},
  booktitle={Proceedings of the IEEE/CVF conference on computer vision and pattern recognition},
  pages={3202--3211},
  year={2022}
}

@inproceedings{he2022masked,
  title={Masked autoencoders are scalable vision learners},
  author={He, Kaiming and Chen, Xinlei and Xie, Saining and Li, Yanghao and Doll{\'a}r, Piotr and Girshick, Ross},
  booktitle={Proceedings of the IEEE/CVF conference on computer vision and pattern recognition},
  pages={16000--16009},
  year={2022}
}

@book{torquato2002random,
  title={Random heterogeneous materials: microstructure and macroscopic properties},
  author={Torquato, Salvatore and others},
  volume={16},
  year={2002},
  publisher={Springer}
}

@article{gostick2019porespy,
  title={PoreSpy: A python toolkit for quantitative analysis of porous media images},
  author={Gostick, Jeff T and Khan, Zohaib A and Tranter, Thomas G and Kok, Matthew DR and Agnaou, Mehrez and Sadeghi, Mohammadamin and Jervis, Rhodri},
  journal={Journal of Open Source Software},
  volume={4},
  number={37},
  pages={1296},
  year={2019}
}

@article{ladd1994numerical,
  title={Numerical simulations of particulate suspensions via a discretized Boltzmann equation. Part 1. Theoretical foundation},
  author={Ladd, Anthony JC},
  journal={Journal of fluid mechanics},
  volume={271},
  pages={285--309},
  year={1994},
  publisher={Cambridge University Press}
}

@article{zou1997pressure,
  title={On pressure and velocity boundary conditions for the lattice Boltzmann BGK model},
  author={Zou, Qisu and He, Xiaoyi},
  journal={Physics of fluids},
  volume={9},
  number={6},
  pages={1591--1598},
  year={1997},
  publisher={American Institute of Physics}
}

@article{lee2024microstructure,
  title={Microstructure reconstruction using diffusion-based generative models},
  author={Lee, Kang-Hyun and Yun, Gun Jin},
  journal={Mechanics of Advanced Materials and Structures},
  volume={31},
  number={18},
  pages={4443--4461},
  year={2024},
  publisher={Taylor \& Francis}
}

@misc{neumann2020,
  author = {Neumann, R. and Andreeta, M. and Lucas-Oliveira, E.},
  title = {{11 Sandstones: raw, filtered and segmented data}},
  year = {2020},
  publisher = {Digital Rocks Portal},
  doi = {10.17612/F4H1-W124},
  url = {https://doi.org/10.17612/F4H1-W124},
  note = {Project DRP-317}
}

@misc{ImperialData,
  title = {Micro-CT Images and Networks},
  author = {{Imperial College London}},
  year = {2015},
  howpublished = {\url{https://www.imperial.ac.uk/earth-science/research/research-groups/pore-scale-modelling/micro-ct-images-and-networks/}},
  note = {Accessed: 2026-01-23}
}

@article{dong2009pore,
  title={Pore-network extraction from micro-computerized-tomography images},
  author={Dong, Hu and Blunt, Martin J},
  journal={Physical Review E},
  volume={80},
  number={3},
  pages={036307},
  year={2009},
  publisher={APS}
}

@article{hu2025pore,
  title={Pore-scale simulation of counter-current spontaneous imbibition in natural fractured porous media},
  author={Hu, Yingxue and Xu, Yusong and Dong, Kai and Huang, Guoqiang and Cai, Meng and Wang, Qingguo and Gu, Zhaolin and Su, Junwei},
  journal={Physics of Fluids},
  volume={37},
  number={8},
  year={2025},
  publisher={AIP Publishing}
}

@article{liu2025pore,
  title={The pore-network-continuum hybrid modeling of nonlinear shale gas flow in digital rocks of organic matter},
  author={Liu, Dongchen and Yang, Xuefeng and Zhang, Deliang and Huang, Shan and Jiang, Rui and Rong, Jianqi and Wang, Zhiwei and Shi, Bowen and Qin, Chao-Zhong},
  journal={Physics of Fluids},
  volume={37},
  number={6},
  year={2025},
  publisher={AIP Publishing}
}

@article{lopes2025enabling,
  title={Enabling FEM-based absolute permeability estimation in giga-voxel porous media with a single GPU},
  author={Lopes, Pedro Cortez Fetter and Semeraro, Federico and Pereira, Andr{\'e} Mau{\'e}s Brabo and Leiderman, Ricardo},
  journal={Computer Methods in Applied Mechanics and Engineering},
  volume={434},
  pages={117559},
  year={2025},
  publisher={Elsevier}
}

@article{zhu2025data,
  title={Data-driven multiscale lattice discrete particle model for digital twin modeling of concrete structures},
  author={Zhu, Yingbo and Brigham, John and Fascetti, Alessandro},
  journal={Computer Methods in Applied Mechanics and Engineering},
  volume={445},
  pages={118183},
  year={2025},
  publisher={Elsevier}
}

@article{ge20263d,
  title={3D microstructure reconstruction of heterogeneous material from slice descriptors using explicit neural network},
  author={Ge, Xiangyun and Wang, Liyuan and Garcia, Liam J and Zhong, Shan and Chen, Bingbing and Li, Chenfeng},
  journal={Computer Methods in Applied Mechanics and Engineering},
  volume={448},
  pages={118469},
  year={2026},
  publisher={Elsevier}
}

@article{chen2025novel,
  title={A novel data-driven digital reconstruction method for polycrystalline microstructures},
  author={Chen, Bingbing and Li, Dongfeng and Wang, Liyuan and Ge, Xiangyun and Li, Chenfeng},
  journal={Computer Methods in Applied Mechanics and Engineering},
  volume={441},
  pages={117980},
  year={2025},
  publisher={Elsevier}
}

\end{document}